\documentclass[twoside]{report}
\usepackage[utf8]{inputenc}

\usepackage{appendix}
\usepackage[top=3cm, bottom=2.85cm, left=2.8cm, right=2.8cm]{geometry}
\usepackage{amsthm}
\usepackage{color}
\usepackage[table,xcdraw]{xcolor}
\usepackage{amsmath,amssymb,amsthm,mathrsfs,amsfonts,dsfont} 

\usepackage[pdftex]{graphicx}
\usepackage{tikz}
\usepackage{fancyhdr}
\usepackage{multirow}
\usepackage{enumitem}
\usepackage{url}

\usepackage{titlesec} 

\usepackage{hyperref}
\hypersetup{
	colorlinks,
    citecolor=black,
    filecolor=black,
    linkcolor=black,
    urlcolor=black}
\usepackage{bbm}	
\usepackage{array} 
\newcolumntype{P}[1]{>{\centering\arraybackslash}p{#1}} 
\newcolumntype{M}[1]{>{\centering\arraybackslash}m{#1}}
\usepackage{textcomp}

\title{Reinforcement Learning for Physical Layer Communications}

\date{\today}
\author{Philippe Mary \\ Visa Koivunen \\ Christophe Moy}
\makeatletter
\let\thetitle\@title
\let\theauthor\@author
\let\thedate\@date
\makeatother

\theoremstyle{plain}

\newtheorem{remark}{Remark}
\newcommand{\set}[1]{\mathcal #1} 
\renewcommand{\vec}[1]{\boldsymbol{#1}} 
\newcommand{\prob}[2]{\mathbb{P}_{#2}\left[#1\right]} 
\newcommand{\Exp}[2]{\mathbb{E}_{#2}\left[#1\right]} 
\newcommand{\ind}[1]{\mathds{1}\left\{#1\right\}} 
\newcommand{\defeq}{\overset{\bigtriangleup}{=}} 

\makeatletter
\renewcommand{\thechapter}{\arabic{chapter}}
\titleformat{\chapter}[display]
    {\bfseries\Large}
  {\filleft\MakeUppercase{\chaptertitlename} \huge\thechapter}
  {4ex}
  {\titlerule
   \vspace{2ex}%
   \filright}
  [\vspace{2ex}%
\titlerule
\vspace{2ex}
\hfill \normalsize  \itshape \begin{tabular}{ r} \@author\end{tabular}]
\makeatother

\begin{document}

\setcounter{chapter}{8}\chapter[RL for PHY layer communications]{Reinforcement Learning for Physical Layer Communications}
\section{Introduction} \label{sec:intro}

Wireless communication systems have to be designed in order to cope with time-frequency-space varying channel conditions and variety of interference sources. In cellular wireless systems for instance, channel is estimated regularly by mobile terminals and base stations (BS) using dedicated pilot signals. This allows for adapting the transmitters and receivers to the current channel conditions and interference scenario. Powerful adaptive signal processing algorithms have been developed in the past decades in order to cope with the dynamic nature of the wireless channel, e.g. the least mean square and recursive least square algorithms for channel equalization or estimation, the Kalman filtering in multiple-input multiple-output channel matrix and frequency offset tracking. These techniques rely on very well established mathematical models of physical phenomena that allow to derive the optimal processing for a given criterion, e.g. mean square error and assumed noise and interference distribution models.

Any mathematical model trades-off between its complexity and its tractability. A very complete, and hence complex, model may be useless if any insight on the state of the system cannot be drawn easily. For instance, the wireless propagation channel is absolutely deterministic and the signal received at any point of the space at any time can be precisely predicted by the Maxwell equations. However, this would require a prohibitive amount of computation and memory storage for a receiver to calculate at any point the value of the electric and magnetic fields using  detailed and explicit knowledge of the physical characteristics of scatterers in the propagation environment, e.g. the dielectric and permittivity constants of the walls and other obstacles. It is much more efficient to design receivers that perform well in environments that have been stochastically characterized instead of using explicit deterministic model of each particular propagation environment.

Modern and emerging wireless systems are characterized by massive amounts of connected mobile devices, BS, sensors and actuators. Modeling such large scale wireless systems has become a formidable task because of, for example, very small cell sizes, channel aware link adaptation and waveform deployment, diversity techniques and optimization of the use of different degrees of freedom in tranceivers. Consequently, it may not be feasible to build explicit and detailed mathematical models of wireless systems and their operational environments. In fact, there is a serious modeling deficit that calls for creating awareness of the operational wireless environment through sensing and learning.     

Machine learning (ML) refers to a large class of algorithms that aim at giving to a machine the capability to acquire knowledge or behavior. If the machine is a wireless system, which is man-made, then the goal of ML in that case is to let the system choose its signal processing techniques and protocols to perform communication without being programmed \emph{a priori}. The learning can be \emph{supervised}, requiring labeled data, \emph{unsupervised} or based on \emph{reinforcement learning} requiring trial-and-error approach. Supervised learning refers to methods that learn from a training set for which the desired output is provided by an external supervisor, i.e. with labeled data, and then perform a task on data that are not present in the training set. Unsupervised learning refers to methods that attempt to find hidden structures or patterns in a large data set. Reinforcement learning (RL), is a class of machine learning technique that aims at maximizing a cumulative \emph{reward} signal over a finite or infinite time horizon for the selected actions. RL refers to the interaction through trials and errors between an \emph{agent}, the entity that learns, and its operational \emph{environment}, and will be the focus of this Chapter. This technique is particularly useful when the agent wants to acquire a knowledge on the characteristics of its environment while making minimal assumptions on it. In this Chapter, we will focus on RL for physical layer (PHY) communications. The rewards in learning are then typically associated with high data rate or high signal-to-interference and noise ratio (SINR). Even if ML in general and RL in particular are what we may call \emph{data driven} approaches, mathematical and physics-based modeling should not be completely abandoned. Indeed, wireless systems are man-made with plenty of structures that can be exploited to make the learning faster, transparent, and explainable. 
   

Three main reasons for RL be applied to a wireless communication problem are: i) the mathematical modelling of the environment is far too complex to be implemented in an agent, or such a mathematical model does not exist, ii) the signalling for acquiring the useful data needed to properly run the system is too complex and iii) the desired outcome or goal of the learning can be described as a scalar reward. For instance, the massive machine type communication or small-cell deployment where detailed network planning is not feasible are interesting scenarios where RL would be attractive technology. Moreover, machine type communications typically involve some cheap battery operated devices where complex base band processing is not possible which is in line with condition i). Moreover, excessive signalling is excluded because of the huge number of devices to be connected, condition ii). Hence, transmission strategies, e.g. channel, transmit power, beam patterns, can be \emph{learned} from scratch or at least with very few \emph{a priori} information in order to maximize the number of received packets, condition iii).

\emph{Adaptability} is the key benefit of RL techniques, as well as classical adaptive signal processing techniques such as the least mean square filter. However, unlike adaptive filtering, RL does not rely on well established mathematical models of the physical phenomena that occur during a wireless transmission or very few (cf. Section \ref{sec:rltheory}). RL techniques select suitable actions based on the feedback received from the surrounding environment in order to maximize a \emph{reward} function over time. An important example of RL applied to physical layer communication is the flexible and opportunistic use of the underutilized frequency bands. 

RL techniques involve a set of possible actions, a set of states and a reward signal. These notions will be rigorously defined in the next section. Some examples of typical actions are transmit on a given band or not, or select among the pre-designed beamformers or pre-coders.  The states of the environment the agent observes are, for instance, whether the selected band is idle or not, or feedback from the receiver that the observed SINR value is below a threshold on certain channels or the transmission has been successful (+1) or not (0 or -1), respectively.

In the particular example of the opportunistic spectrum access, the cardinality of action and state sets is small. Hence the learning or convergence phase is faster than in a scenario where the action space would be much larger. If we imagine a smart radio device that is able to do link adaptation by choosing the modulation technique, transmit power and to select a channel to transmit its signal, the search space would become much larger. Hence, the convergence towards the optimal result that maximizes a given reward function, would take much longer time and may become complex for RL approaches. However, recent advances in Deep RL methods have alleviated this problem. An example on this scenario, i.e. link adaptation, will be given in this chapter. Another practical example considered later in this chapter is smart radio access network with the goal to optimize the energy consumption of a cluster of BSs. This is a demanding problem with a high dimensional state space which makes finding the optimal solution using classical optimization technique difficult. RL may help to find a good solution without completely avoiding the problem complexity, though.


Deep learning (DL) strategies involve multiple layers of artificial neural networks (ANN) that are able to extract hidden structures of labeled (\emph{supervised learning}) or unlabeled data (\emph{unsupervised learning}) \cite{Goodfellow2016}. These techniques prove to be very efficient in many applications such as image classification, speech recognition and synthesis, and all applications involving a large amount of data to be proceed. Recently, researchers in wireless communications have shown interest for using ML and DL in wireless networks design (see \cite{Zappone2019,Chen2019} and references therein for a complete state of the art on that topic), mainly for networking issues, but more recently also for PHY layer, such as user-channel matching or performance optimization in large scale randomly deployed networks. This is particularly useful when a complete mathematical model of the behavior of a network is intractable due for example to the large dimensionality of the state or action spaces. RL and DL can be combined into \emph{deep reinforcement learning} (DRL) approach when the number of observable states or possible actions are too large for conventional RL. This technique relies on the reinforcement learning principle, i.e. an agent interacting with its environment, but the action to be taken is obtained through a non-linear processing involving neural networks (NN) \cite{Luong2019}.


In this chapter, we will give comprehensive examples of applying RL in optimizing the physical layer of wireless communications by defining different class of problems and the possible solutions to handle them. In Section \ref{sec:rltheory}, we present all the basic theory needed to address a RL problem, i.e. Markov decision process (MDP), Partially observable Markov decision process (POMDP), but also two very important and widely used algorithms for RL, i.e. the Q-learning and SARSA algorithms. We also introduce the deep reinforcement learning (DRL) paradigm and the section ends with an introduction to the multi-armed bandits (MAB) framework. Section \ref{sec:apps} focuses on some toy examples to illustrate how the basic concepts of RL are employed in communication systems. We present  applications extracted from literature with simplified system models using similar notation as in Section \ref{sec:rltheory} of this Chapter. In Section \ref{sec:apps}, we also focus on modeling RL problems, i.e. how action and state spaces and rewards are chosen. The Chapter is concluded in Section \ref{sec:trends} with a prospective thought on RL trends and it ends with a review of a broader state of the art in Section \ref{sec:relatedworks}.

\paragraph{Notations:} The table at beginning of the Chapter summarizes the notation used in the following. We review the main ones here. Random variables, and stochastic processes when depending on time, are denoted in capital font, e.g. $X(t)$, while their realizations in normal font, i.e. $x(t)$. Random vectors are denoted in capital bold font, e.g. $\mathbf{X}(t)$, and their realizations in small bold font, i.e. $\mathbf{x}(t)$. The conditional probability distribution of a random variable $X$ at time $t+1$ given another random variable $Y$ at $t$, is denoted as $P_{X(t+1)\left|\right.Y(t)}(x\left|\right. y)$ and when no ambiguity is possible on the underlying distribution function, simply by $ p(x\left|\right. y)$, being understood that the first and the second arguments in $p\left(\cdot\left|\right. \cdot\right)$ rely to the values of the random variables at time $t+1$ and $t$ respectively, except otherwise mentioned. Vectors and matrices are denoted with bold font. Moreover, $\ind{a}$ is the indicator function, which is equal to 1 if $a$ is true and 0 otherwise. $D\left(P\left|\right|Q\right)$ is the Kullback-Leibler divergence between the distributions $P$ and $Q$ and is defined as
$$
D\left(P\left|\right|Q\right) = \int_{\set{I}} \log \frac{dP}{dQ} dP \defeq \Exp{\log \frac{dP}{dQ}}{P} ,
$$
where $dP/dQ$ is called the Radon-Nikodym derivative and is defined if $P\ll Q$, i.e. the measure $P$ is absolutely continuous w.r.t. the measure $Q$. This means that for all $x \in \set{I}$, if $Q(x) = 0$ then $P(x) = 0$. This definition holds if the probability measures are continuous but also if they are discrete. In the former case, the Radon-Nikodym derivative is simply the ratio between the density probability functions and in the latter case, it is the ratio of the probability mass functions that can be denoted as $p$ and $q$.

\section{Reinforcement Learning: background} \label{sec:rltheory}

\subsection{Overview}

RL may be considered as a sequential decision making process. Unlike supervised learning it does not need annotation or labeling of input-output pairs. The decision-maker in RL is called an agent.  An agent has to sequentially make decisions that affect future decisions.  The agent interacts with its environment by taking different actions. There are a number of different alternative actions the agent has to choose from. Selecting the best action requires considering not only immediate but also long-term effects of its actions. After taking an action, the agent observes the environment state and receives a reward.  Given the new state of the environment, the agent takes the next action. A trajectory or history of subsequent states, actions and rewards is created over time. The reward quantifies the quality of the action and consequently captures what is important for the learning system. The objective of the learning is typically to maximize the sum of future rewards but also some other objective that is a function of the rewards may be employed. The learning takes place by trial and error so that the rewards reinforce decisions that move the objective towards its optimum. As a consequence the system learns to make good decisions and can independently improve its performance  by probing different actions for different states. Hence, consequences to future actions are learned. Commonly the immediate rewards are emphasized and the future rewards are discounted over time. Sometimes actions with small immediate reward can lead to a higher payoff in a long term. One would like to choose the action that trades off between the immediate rewards and the future payoffs the best possible way. 

RL are typically modeled using an MDP framework. The MDP provides a formal model to design and analyze RL problems as well as a rigorous way to design algorithms that can perform optimal decision making in sequential scenarios. If one models the problem as an MDP, then there exists a number of algorithms that will be able to automatically solve the decision problem. However, real practical problems cannot be strictly modeled with an MDP in general, but this framework can be a good approximation of the physical phenomena that occur in the problem such that this model may perform well in practice.

\subsection{Markov Decision Process} \label{subsec:mdp}

An MDP model is comprised of four components: a set of states, a set of actions, the transitions, i.e. how actions change the state and the immediate reward signal due to the actions. We consider a sequence of discrete time steps $t= \{0, 1, 2,...\}$ and we briefly describe each component. The state describes the set of all possible values of dynamic information relevant to the decision process. In a broad sense, one could think that a state describes the way the world currently exists and how an action may change the state of the world. Obviously we are considering only an abstract and narrow view of the world that is relevant to the learning task at hand and our operational environment. The state at time instance $t$ is represented by a random variable $S(t)\in \set{S}$ where $\set{S}$ is the state space, i.e. the set of all possible values of dynamic information relevant to the learning process. In the context of physical layer of wireless communications, the state could describe, for example, the occupancy of the radio spectrum, or the battery charging level of the user equipment. 

The environment in which the agent acts is modeled as a Markov chain. A Markov chain is a stochastic model for describing the sequential state transitions in memoryless systems. A memoryless transition means that
\begin{equation} \label{eq:memlessdef}
    P_{S(t+1)\left|\right. \vec{S}(t)}\left(s' \left|\right. \vec{s}\right) = P_{S(t+1)\left|\right. S(t)}\left(s' \left|\right. s \right) ,
\end{equation}
with
\begin{equation}
    P_{S(t+1)\left|\right. S(t)}\left(s' \left|\right. s \right) \defeq \prob{ S(t+1)=s'\left|\right. S(t)=s}{},
\end{equation}
if $\set{S}$ is a discrete set. Moreover $\vec{S}(t) = \left[S(0), \cdots, S(t) \right]$ and $\vec{s} = [s(0), \cdots, s(t)]$, $S(t)$ is the state at time instance $t$ and $P_{S(t+1)\left|\right.S(t)}$ is the conditional distribution of $S(t+1)$ given $S(t)$. In other words, the probability of the state transition to state $S(t+1)$ is only dependent on the current state $S(t)$. Hence, there is no need to remember the history of the past states to determine the current state transition probability.

The actions are the set of possible alternatives we can take. The problem is to know which of these actions to choose in a particular state of the environment. The possible actions the agent can choose at state $S(t)$ is represented with the random variable $A(t) \in \set{A}$ where $\set{A}$ is the action space. The possible actions may depend on the current state. Typical actions in a wireless communication context can be giving access to the channel, sensing the channel, selecting a beam pattern or adjusting the power, for instance.

The action the agent takes activates a state transition of the Markov chain.
The state transition function specifies in a probabilistic sense the next state of the environment as a function of its current state and the action the agent selected.  It defines how the available actions may change the state. Since an action could have different effects, depending upon the state, we need to specify the action's effect for each state in the MDP. The state transition probabilities depend on the action such as
\begin{equation} \label{eq:statetrans}
P_{S(t+1)| S(t),A(t)}(s'\left|\right. s, a) \defeq p(s'\left|\right. s, a).
\end{equation}
If we want to make the decision making process automatic and run it in a computer, then we must be able to quantify the value associated to the actions taken. The value is a scalar and called a reward. The reward from action $A(t)$ is represented by a random variable $R(t) \in \set{R}$, where $\set{R}$ is the set of rewards, and it specifies the immediate value for performing each action in each state. 

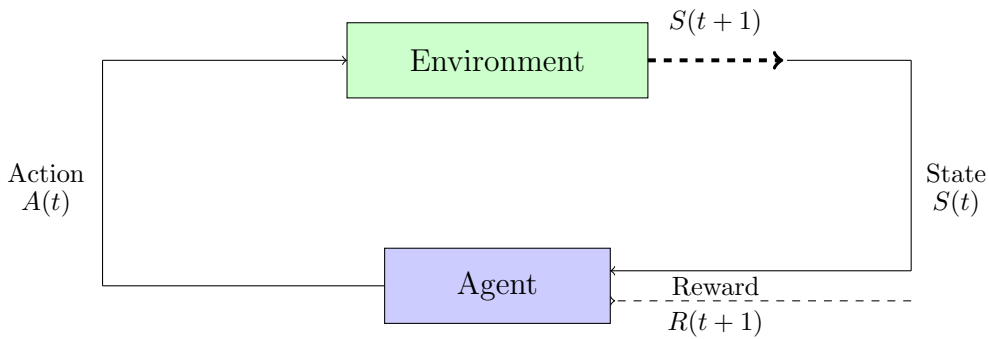
\begin{figure}[htbp]
\begin{center}
    \begin{tikzpicture}
        \draw[fill=green!20] (-2,0) rectangle (2,1) ; 
        \node at(0,0.5){{\large Environment}}; 
        \draw[fill=blue!20] (-1.5,-3) rectangle (1.5,-2); 
        \node at(0,-2.5){{\large Agent}}; 
        \draw[dashed, line width=1.5pt, ->] (2,0.5) -- (3.8,0.5); 
        \draw (3.85,0.5) -- (5.5,0.5); 
        \draw (5.5,0.5) -- (5.5, -2.3); 
        \draw[->] (5.5,-2.3) -- (1.5,-2.3); 
        \draw[dashed, -<] (5.5,-2.7) -- (1.5,-2.7) ; 
        \draw (-1.5,-2.5) -- (-5.25,-2.5); 
        \draw (-5.25,-2.5) -- (-5.25,0.5); 
        \draw[->] (-5.25,0.5) -- (-2,0.5); 
        \node at(-6,-1){Action}; 
        \draw (-6,-1.1) node[below]{$A(t)$};
        \node at(6.1,-1){State}; 
        \draw (6.1,-1.1) node[below]{$S(t)$};
        \draw (2.9,0.7) node[above]{$S(t+1)$}; 
        \node at(2.9,-2.5){Reward} ;
        \draw (2.9,-2.7) node[below]{$R(t+1)$}; 
    \end{tikzpicture}
\end{center}
\caption{Reinforcement learning principle in MDP} \label{fig:rl}
\end{figure}

The reward provides the means for comparing different actions and choosing the right one for achieving the objective of the learning process. The reward distribution depends on which action was selected and on the state transition. The conditional probability distribution defining the dynamics of an MDP process is
\begin{equation} \label{eq:srdist}
    P_{R(t+1) S(t+1)\left|\right. S(t) A(t)}\left(r, s' \left|\right. s, a \right) \defeq p(r,s'\left|\right. s,a).
\end{equation}
An MDP can last for a finite or infinite number of time steps, distinguishing between finite and infinite horizon models and methods. The interconnection between the different elements of an MDP can be illustrated with the well-known diagram of Fig.~\ref{fig:rl}.

A reward function $R()$ gives a payoff $R(t+1)$ for choosing action $A(t)$ in state $S(t)$ resulting in new state $S(t+1)$. After selecting an action the agent obtains the reward and the next state, but no information on which action would have been the best choice towards its objective in a long term. Hence, the agent will perform active probing and obtain experience about the possible system states, available actions, transitions and rewards to learn how to act optimally. A very widely used objective is to maximize the expected sum of discounted rewards over an infinite horizon \cite{Sutton2018,Szepesvari2010}:
\begin{equation} \label{eq:discreward}
J_{\pi} = \Exp{ \sum_{t=0}^{\infty} \gamma^t R(t+1) }{\pi} ,
\end{equation}
where $\gamma<1$ is the discount factor emphasizing more immediate rewards and the expectation is taken over the distribution of the rewards, following a certain \emph{policy} $\pi$ whose the meaning will be detailed hereafter. Discounting is the most analytically tractable and hence the most commonly used approach. Other objective functions may be employed including expected reward over a finite time horizon and regret, which measures the expected loss in the learning compared to an optimal policy. 

The solution to an MDP is called a policy and it simply specifies the best sequence of actions to take during the learning process.  Policy maps states to actions, i.e. $\pi: \set{S}\rightarrow \set{A}$ and can be \emph{deterministic}, i.e. a single or a set of deterministic actions is performed when the agent encounters the state $S(t)$, or it can be \emph{stochastic} and is defined as the conditional probability measure of $A(t)$ given $S(t)$ i.e. $P_{A(t)\left|\right.S(t)}\left(a\left|\right.s\right)$, often simply denoted as $\pi(a\left|\right.s)$\footnote{Note in that case, both $a$ and $s$ refers to the value of the action and the state respectively at time $t$, because action is immediately chosen while observing a given state and not delayed to the next time slot.}. It is basically a sequence of the decision rules to be used at each decision time instance. For the infinite-horizon discounted model, there exists an optimal deterministic stationary policy. The finite-horizon model is appropriate when the system has a hard deadline or the agent's lifespan is known. 

An optimal policy would need to consider all the future actions. Such policies are called non-myopic policies. In comparison, a policy where the agent maximizes the immediate reward is called a myopic policy. Typically, the myopic policies are suboptimal for $\gamma > 0$. The goal is to derive a policy which gives the best actions to take for each state, for the considered horizon.

\subsubsection{Value functions}

In order to be able to derive the optimal policy that maximizes the function $J$ in \eqref{eq:discreward}, one needs to evaluate the value of the states under a certain policy $\pi$, i.e. a function $v_{\pi} : \set{S}\rightarrow \mathbb{R}$. This is defined as the expectation, over the policy $\pi$, of the discounted reward obtained when starting from the state $s\in \set{S}$, that is to say \cite{Sutton2018}
\begin{equation} \label{eq:statevalue}
    v_{\pi}(s) = \Exp{ \sum_{t=0}^{\infty} \gamma^t R(t+1) \left|\right. S(0) = s }{\pi}, \forall s\in \set{S} .
\end{equation}
Since the state value function depends on the state $s$ taken for the computation, this quantity may be averaged over the states in order to obtained the average reward in \eqref{eq:discreward}. The state value function represents the average of the discounted reward that would be obtained starting from an initial state $s$ and following the policy $\pi$.

Similarly, one can defined the \emph{action-state} function, $q_{\pi}: \set{S}\times \set{A}\rightarrow \mathbb{R}$, by averaging the discounted reward when starting at $t=0$ from state $s\in \set{S}$ and taken the action $a\in \set{A}$, and then following the policy $\pi$ thereafter. We hence have now an expectation conditioned on the state and the action, that is \cite{Sutton2018,Szepesvari2010}
\begin{equation} \label{eq:actionstatevalue}
    q_{\pi}(s,a) = \Exp{\sum_{t=0}^{\infty} \gamma^t R(t+1) \left|\right. S(0) = s, A(0) = a }{\pi}, \forall (s,a) \in \set{S}\times \set{A} .
\end{equation}

\subsubsection{Bellman equations} \label{sec:bellman}
A remarkable property of the state value function in \eqref{eq:statevalue} is that it follows a recursive relation that is widely known as the Bellman equation \cite{Bellman1957} of the state value function. For all $s\in \mathcal{S}$, one can prove \cite{Sutton2018}:
\begin{equation}
    v_{\pi} (s) = \sum_{a \in \set{A}} \pi(a\left|\right. s) \sum_{\left(s',r\right) \in \set{S}\times \set{R}} p(s',r\left|\right. s,a) \left[r + \gamma v_{\pi}(s')\right] \label{eq:bellmanfinal} .
\end{equation}
The expression in \eqref{eq:bellmanfinal} is known as the Bellman's equation for the state value function and has to be seen as the expectation of the random variable $R(t+1) + \gamma v_{\pi}(S(t+1))$ over the joint distribution $P_{A(t)\left|\right. S(t)} P_{S(t+1)R(t+1)\left|\right. S(t)A(t)}$. The Bellman's equation links the state value function at the current state with the next state value function averaged over all possible states and rewards knowing the current state and the policy $\pi$. The value of a state $S(t)$ is the expected instantaneous reward added to the expected discounted value of the next states, when the policy $\pi$ is followed. The proof of this relation relies on separating \eqref{eq:statevalue} in two terms, i.e. $t=0$ and $t>0$, and using essentially the Markovian property and the Bayes' rule in the second term to make appear $v_{\pi}(s')$.

The optimal state value and action-state value functions are obtained by maximizing $v_{\pi}(s)$ and $q_{\pi}(s,a)$ over the policies, that is \cite{Szepesvari2010}
\begin{equation}\label{eq:optiv}
    v_*(s) = \sup_{\pi\in \Phi} v_{\pi}(s) ,
\end{equation}
and
\begin{equation}\label{eq:optiq}
    q_*(s,a) = \sup_{\pi\in \Phi} q_{\pi}(s,a) ,
\end{equation}
where $\Phi$ is the set of all stationary policies, i.e. policies that do not evolve with time\footnote{Either a fixed over time \emph{deterministic} policy or a stationary \emph{stochastic} policy, in the strict sense, i.e. the conditional law of $A(t)$ given $S(t)$ does not depend on $t$.}. Moreover, \eqref{eq:optiv} can be written w.r.t. \eqref{eq:optiq} as $v_*(s) = \sup_{a\in \set{A}} q_*(s,a)$. The optimal state value function also obeys to the Bellman recursion and one can show that \cite{Sutton2018,Szepesvari2010}
\begin{equation} \label{eq:bellmanoptiv}
    v_*(s) = \sup_{a\in A} \left\{ \sum_{(r,s')\in \set{S}\times \set{R}} p(r,s'\left|\right. s,a) \left[r + \gamma v_*(s')\right]\right\} \mbox{ } \forall s\in \set{S} .
\end{equation}
This last equation means that the expected return from a given state $s$ and following the optimal policy $\pi^*$ is equal to the expected return following the best action from that state. Moreover, substituting $v_*(s')$ in \eqref{eq:bellmanoptiv} with the supremum over the actions of $q_*(s',a)$, we obtain the iterative property on the action-state value function:
\begin{equation} \label{eq:bellmanoptiq}
    q_{*}(s,a) = \sum_{(s',r) \in \set{S}\times \set{R}} p(s',r \left|\right. s,a) \left[ r + \gamma \sup_{a'\in \set{A}} q_*(s',a') \right] .
\end{equation}
Expressions in \eqref{eq:bellmanoptiv} and \eqref{eq:bellmanoptiq} are the Bellman optimality equations for $v_*$ and $q_*$ respectively.


The policy should mostly select the actions that maximize $q_*(s,\cdot)$. A policy that chooses only actions that maximize a given action-state function $q(s,\cdot)$ for all $s$
is called \emph{greedy} w.r.t. $q$. A greedy policy is a decision procedure based on the immediate reward without considering the long term payoff. In general, only considering local or immediate reward may prevent from finding the highest payoff on the long term. However, since $v_*$ or $q_*$ already contain the reward consequences of all possible states, then a greedy policy w.r.t. $q_*$ is hence optimal on the long run \cite{Szepesvari2010}. Hence, if one is able to compute $q_*$, the optimal policy follows. 

\subsubsection{Policy evaluation}

RL algorithms aim at finding the best sequences of actions in order to get close to  \eqref{eq:bellmanoptiv}. As we will see later with the Q-learning, the idea is to keep an estimate of the optimal state or action state value functions at each time step and find a way to make them converge to the optimal value functions. The expressions in \eqref{eq:bellmanoptiv} and \eqref{eq:bellmanoptiq} are fixed point equations and can be solved using dynamic programming. Given a policy $\pi$, the policy evaluation algorithm computes firstly finds the value function for immediate rewards and then extends the time horizon one by one. Basically one just adds the immediate reward of each of the available actions to the value function computed in previous step. This way one builds the value function in each iteration based on the previous one. Let consider a deterministic policy $a = \pi(s)$. A sequence of value functions can be obtained for all states $s$ such as
\begin{equation}\label{eq:itev}
    v_{t+1}(s) = \underbrace{r(s,a)}_{\text{immediate average reward}} + \gamma\underbrace{\sum_{s' \in\set{S}} p(s'\left|\right. s, a) v_t(s'),}_{\text{value function of previous steps}}
\end{equation}
where the immediate average reward, $r(s,a)$, is the expectation over the distribution of the rewards knowing the action $\pi(s)$ taken in the state $s$ at time $t$, i.e. $r(s,a) = \Exp{R(t+1)\left|\right. S(t) = s, A(t) = a}{P_{R\left|\right. S A}}$. This iterative equation can be computed for each action taken in each state. This equation converges because the linear operator linking the value function of state $S(t)$ to the value functions of states $S(t+1)$ is a maximum norm contraction \cite{Szepesvari2010}.

Similarly, the action-state value iteration algorithm aims at iteratively solving the following equation for a deterministic policy
\begin{equation}\label{eq:iteq}
    q_{t+1}(s,a) = \underbrace{r(s,a)}_{\text{immediate average reward}} + \gamma\underbrace{\sum_{s' \in\set{S}} p(s'\left|\right. s, a) q_t(s',a).}_{\text{action-state value function of previous steps}}
\end{equation}
The problem of this procedure is that the agent needs to know the complete dynamic of the MDP, i.e. $p(r,s'\left|\right. s, a)$, that is rarely the case in practice. Fortunately, there are algorithms like Q-learning that converge to the optimal policy without knowing the dynamic of the physical process.

\subsubsection{Policy improvement} \label{sec:policyite}
The iterative procedures described above allow us to evaluate the performance of a policy. Indeed, by choosing a given policy $\pi_0$ one can compute \eqref{eq:itev} or \eqref{eq:iteq} until convergence, e.g. $\left| v_{t+1}(s) - v_t(s)\right|\leq \epsilon$, where $\epsilon$ is an arbitrary small number. When the procedure has converged, the state value obtained, i.e. $v_{\pi_0}(s)$ is the state value function achieved by following the policy $\pi_0$ for state $s$. Hence, one has the \emph{evaluation} of the performance of the policy $\pi_0$. But we still do not know if it is the best policy, i.e. leading to the maximal value of the state function $v_*(s)$. 

In order to progress to the optimal policy, one needs a \emph{policy improvement} step. The idea is to create a new policy $\pi'$ that differs from $\pi$ by a different action taken when being in state $s$, e.g. $\pi'(s) \neq \pi(s)$. If we are able to choose an action when being in state $s$ such that $q_{\pi}(s,\pi'(s)) \geq v_{\pi}(s)$, then $v_{\pi'} (s)\geq v_{\pi}(s)$,  $\forall s\in \set{S}$ \cite{Sutton2018}. Basically, it means that by taking an action $a'$ such that $a'\neq \pi(s)$ and following the policy $\pi$ hereafter such that $q_{\pi}(s,a') \geq v_{\pi}(s)$, the policy $\pi'$ should not be worse than $\pi$ on the performance of the learning task. Let $\pi_t$ denote the policy obtained at iteration $t$. Once \eqref{eq:iteq} has converged under $\pi_t$, we create $\pi_{t+1}$ by choosing the greedy policy w.r.t. $q_{\pi_t}$ which is obtained by choosing the action that maximizes the right-hand side of \eqref{eq:iteq} for all states $s$, and we repeat the process until having no policy improvements.

\subsubsection{Combination of policy evaluation and improvement}

In practice, the two procedures introduced above, i.e. the policy evaluation and improvement, can be combined in a one algorithm called: \emph{value iteration}. This algorithm consists in taking the action that maximizes the (action-)state value function at each time step, i.e. taking the supremum over $\mathcal{A}$ of the right hand side of \eqref{eq:itev} for the state value function and replacing $q_t(s',a)$ by $\sup_{a'\in \mathcal{A}} q_t(s',a')$ in \eqref{eq:iteq} for the action-state value function. The new equations obtained also converge to $v_*$ and $q_*$ thanks to the fixed-point theorem and the property of the Bellman equation. It means that a greedy policy w.r.t. $v_t$ ($q_t$) allows to converge to the optimum (action-)state value.


\subsection{Partially observable Markov decision process}

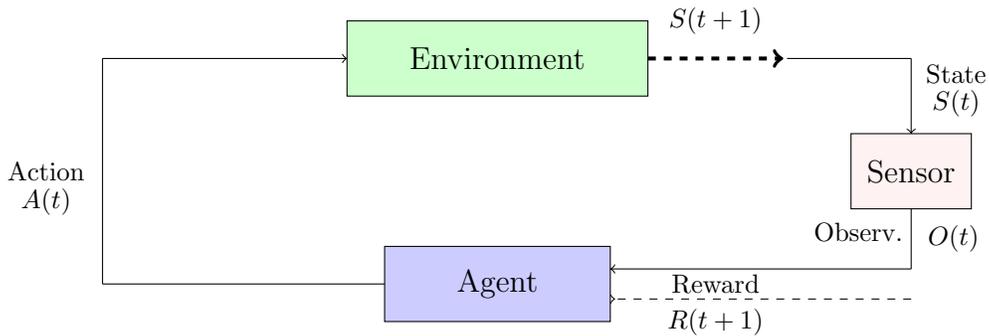
\begin{figure}[htbp]
\begin{center}
    \begin{tikzpicture}
        \draw[fill=green!20] (-2,0) rectangle (2,1) ; 
        \node at(0,0.5){{\large Environment}}; 
        \draw[fill=blue!20] (-1.5,-3) rectangle (1.5,-2); 
        \node at(0,-2.5){{\large Agent}}; 
        \draw[fill=pink!20] (4.7,-1.5) rectangle (6.3,-0.5); 
        \node at(5.5,-1){{\large Sensor}}; 
        \draw[dashed, line width=1.5pt, ->] (2,0.5) -- (3.8,0.5); 
        \draw (3.85,0.5) -- (5.5,0.5); 
        \draw[->] (5.5,0.5) -- (5.5, -0.5); 
        \draw (5.5,-1.5) -- (5.5, -2.3); 
        \draw[->] (5.5,-2.3) -- (1.5,-2.3); 
        \draw[dashed, -<] (5.5,-2.7) -- (1.5,-2.7) ; 
        \draw (-1.5,-2.5) -- (-5.25,-2.5); 
        \draw (-5.25,-2.5) -- (-5.25,0.5); 
        \draw[->] (-5.25,0.5) -- (-2,0.5); 
        \node at(-6,-1){Action}; 
        \draw (-6,-1.1) node[below]{$A(t)$};
        \node at(6.1,0.3){State}; 
        \draw (6.1,0.2) node[below]{$S(t)$};
        \draw (2.9,0.7) node[above]{$S(t+1)$}; 
        \node at(2.9,-2.5){Reward} ;
        \draw (2.9,-2.7) node[below]{$R(t+1)$}; 
        \node at(4.8,-1.8){Observ.}; 
        \draw (5.6,-1.9) node[right]{$O(t)$};
    \end{tikzpicture}
\end{center}
\caption{Reinforcement learning principle in POMDP. Contrarily to MDP in Fig. \ref{fig:rl} the state of the environment is observed through a sensor that gives a partial state of the environment. For instance, the sensor may be the receiver in PHY layer communication and actions would be adapting the operational parameters of the transmitter/receiver pair.} \label{fig:pomdp}
\end{figure}

Complete observability is required for MDP based learning. When the states of the underlying Markovian process are not directly observable, the situation is referred as a POMDP. This model adds sensor measurements that contain information about the state of the MDP model as well as observation function describing the conditional observation probabilities. Instead of observing the current state of the process directly, we may just have access to a noisy version or estimate of the state. Typically, we use physical sensors to acquire a set of noisy observations containing relevant information about the state. The principle of reinforcement learning under POMDP is illustrated on Fig. \ref{fig:pomdp}. For example, in physical layer wireless communication we need to estimate key system parameters such as the channel impulse or frequency response, channel matrix and its covariance, SINR, perform time and frequency synchronization as well as design transmit or receive beamformers, precoders or decoders from data acquired by the receivers. The received signals contain unobservable random noise and harmful interferences. Moreover, in order to adapt our transmitters and optimize the resources usage, we may need to use feedback from receivers or exploit channel reciprocity that are both subject to error. Most of the parameters characterizing the state of a wireless system at the physical layer are defined in a continuous space.  Hence,  the conventional MDP model is not necessarily applicable. The errors caused by noisy sensors are unobservable and random and need to be described using probability distributions. Consequently, the states are also described in terms of probability distributions.

In MDP-based learning our goal is to find a mapping from states to actions. In case of POMDPs, we are looking for a mapping from probability distributions associated with the states to actions. The probability distribution over all possible model states is called {\em  belief state} in POMDP jargon and the entire probability space (i.e. the set of all possible probability distributions) is called the {\em belief space}. The observations acquired by physical sensors contain unobservable and random noise. Hence, we need to specify a probabilistic observation model called observation function in POMDP jargon. This observation model simply tells us the probability of each observation for each state in the model. It is formally given as a conditional probability of an observation given a state-action couple. As a result, we will have uncertainty about the state due to incomplete or uncertain information. That means we have a set of states, but we can never be certain in which state we are. The uncertainties are handled by associating a probability distribution over the set of possible states. The distribution is defined by the set of observations, observation probabilities, and the underlying MDP. The POMPD model considers {\em beliefs} of the states instead of the actual states in MDP. The belief is a measure in interval [0, 1] that describes the probability of being in a specific state. The agent interacts with the environment and receives observations, and then updates its belief state by updating the probability distribution of the current state.  In practice, the agent updates its belief state by using a state estimator based on the last action, current acquired observation and the  previous belief state.  By solving a POMDP problem we find a policy that tells which action is optimal for each belief state.  

A POMDP is a tuple $<\mathcal{S},\mathcal{A},\mathcal{R},\Omega, O>$. The environment is represented by a set of states $\set{S}= \{s_1,s_2,...,s_N\}$ with $|\set{S}|=N$, a set of possible  actions denoted by $\set{A}= \{a_1,a_2,...,a_K\}$ with $|\set{A}|=K$. The transition from state $s$ to new state $s'$  given the action $a$ is governed by a probabilistic transition function as in \eqref{eq:statetrans} that is the conditional distribution $P_{S(t+1)|S(t), A(t)}$. As an immediate result, the agent receives a reward $R(t+1)$ that depends on the state $s$ the agent was in and the action $a$ it took. These components of the POMDP-tuple are as in a conventional MDP. Since state $s'$ is not directly observable, it can be estimated by using observations that contain information about the state. A finite set of observations emitted by the state is denoted by $\vec{z} \in \Omega$ with $|\Omega|=M$. An observation function defines a probability distribution for each action $A(t)$ and the resulting state $S(t+1)$. The observation $Z(t+1)$ is then a random variable and the probability distribution of the observations conditioned on the new state $S(t+1)$ and action $A(t)$ is $P_{Z(t+1)| S(t+1), A(t)}(z\left|\right. s', a)\defeq O(z, s',a)$. 

In a POMDP, the following cycle of events takes place. An environment is in state $S$ and the agent computed the belief state $B$. The agent takes an action $A$ using the policy $\pi(\cdot\left|\right. B)$ and the environment transitions to a new state $S'$ according to the state transition distribution. The new state is not directly observable, but the environment emits an observation $Z$ according to the conditional distribution $O(z , s',a)$. Then the agent receives a reward $R$ for action taken from belief state $B$. Finally, the agent updates the belief state and runs the  cycle again. As a result, a trajectory or history of subsequent belief states, actions, observation and rewards is created over time. 
The goal of an agent is to learn a policy $\pi$ that finds actions that maximize the policy's value. There are many ways to measure the quality. The most common choice is the discounted sum of rewards as in the case of MDP.

The belief state is a sufficient statistic that contains all the relevant information about the past observations and enjoys the Markov property.  The next belief state depends only on the current belief state and the current action and observation. Hence, there is no need to remember the whole history of actions and observations. Updating the distribution requires using the transition and observation probabilities and a formula stemming from the Bayes rule. Let us denote $P_B(s)\defeq \prob{S=s}{B}$ the distribution of the belief state, which is the probability that the environment is in state $s$ under the distribution of the belief. The belief state update that can be considered as a state estimation step in POMDP framework is given by
\begin{equation}
P_{B}(s') = \frac{O(z , s',a)}{p(z|b,a)} \sum_{s \in \set{S}} p(s'\left|\right. s,a)P_B(s). 
\end{equation}
The denominator is a normalizing constant that makes the sum on $s'$ equal to one, i.e. $p(z|b,a)= \sum_{s' \in \set{S}} O(z , s',a) \sum_{s \in \set{S}} p(s'|s,a) P_B(s)$. The distributions are updated simply by applying the Bayes' Rule and using the model parameters. Similarly to MDP, the agent wishes to choose its actions such that it learns an optimal policy $\pi^*(b)$. 

The main practical difficulty in POMDP models is finding a solution that is a policy that chooses optimal action for each belief state. One may use value iteration or policy iteration approaches as in the case of MDP. Now the state space is just defined in terms of probability distributions. Algorithms used for solving POMDP typically stem from dynamic programming. 
It can be solved using value iteration, policy iteration or a variety of other techniques developed for solving MDP. Optimal value function $v_*$ is then the value function associated with  an optimal policy $\pi^*$. The early approach for solving POMDPs was using belief-MDP formulation. The value iteration follows the Bellman's equation already introduced for MDP, i.e.
\begin{eqnarray}
v_0(b)& = & 0 . \\
v_{t+1}(b) & = & \sup_{a \in \set{A}} [ r(b,a) + \gamma \sum_{z \in \Omega} p(z\left|\right. b,a) v_t(b')],
\end{eqnarray}
where $r(b,a) = \sum_{s \in \set{S}} P_B(s) r(s,a)$ is the average reward for action $a$ in the belief state $b$ and where $r(s,a)$ is the expected reward obtained in state $s$ and taking action $a$ that has been defined in \eqref{eq:itev} and \eqref{eq:iteq}. Moreover, $p(z\left|\right. b,a)$ has been defined previously and denotes the conditional distribution of the observation $z$ at time $t+1$, given the belief state $b$ and taking action $a$ at $t$. A value function may be modeled with a structure that can be exploited in making the problem solving easier or even  feasible. For example a piecewise linear and convex approximation may be used in finite horizon scenarios. Value iteration provides an exact way of determining the value function for POMDP. Hence, the optimal action can be found from the value function for a belief state. Unfortunately, the complexity of solving POMDP problem via value iteration is exponential in the number of observations and actions.  Moreover, the dimensionality of the belief space grows proportionally to the number of states. The dimensionality of belief space can be reduced using a parametric model such as a Gaussian mixture model.  Also point-based value iteration has been proposed for solving POMDP. In such methods, a small set of reachable belief points are selected and the Bellman updates are done at these points while storing values and gradients. Heuristic methods employing search trees have been developed, too. The methods build an AND/OR tree of reachable belief states from the current belief and perform a search over the tree using well-known methods such as branch-and-bound.

Value iteration is used very commonly since the classical policy iteration algorithm for POMDP proposed in \cite{Smallwood1973} has a high complexity and is thus less popular. There are, however, algorithms of lower complexity. One can also use the policy iteration methods as it has been described in Section \ref{sec:policyite} for solving POMDP problems.

\subsection{Q-learning and SARSA algorithm} \label{sec:qsarsa}

Q-learning and the current State, current Action, next Reward, next State and next Action (SARSA) algorithms are iterative procedures to learn the optimal policy that maximize the expected reward from any starting state, and without the knowledge of the MPD dynamics. Both algorithms exhibit some similarities but differ in some key points that we detail hereafter. Q-learning and SARSA are called tabulated algorithms, i.e. they are based on the construction of a look-up table, a.k.a Q-table, that allows the algorithm to trace and update the expected action-value function for all action-state pairs $(s,a)\in \set{S}\times \set{A}$. Once the Q-table has been built, the optimal policy is to choose the action with the highest score. The basic idea of both algorithms is to build a new estimate from an old estimate, which is updated by an incremental difference between a target and the old estimate. This can be formalized as follows:
\begin{equation} \label{eq:qlearning}
    \underbrace{q_t\left(S(t),A(t)\right)}_{\text{new estimate}} \leftarrow \underbrace{q_t\left(S(t),A(t)\right)}_{\text{old estimate}} + \alpha_t \left[ \underbrace{T_{t+1}}_{\text{target}} - \underbrace{q_t\left(S(t),A(t)\right)}_{\text{old estimate}}\right] ,
\end{equation}
where $\alpha_t$ is the \emph{learning rate} at time $t$ which is a scalar between 0 and 1. The learning rate tells us how much we want to explore something new and how much we want to exploit the current choice. Indeed, in each of the reinforcement learning algorithms, exploitation and exploration have to be balanced in order to trade-off between the desire for the agent to increase its immediate reward, i.e. exploiting actions that gave good results so far, and the need to explore new combinations to discover strategies that may lead to larger rewards in the future, this is the exploration. At the beginning of the procedure, one may expect to spend more time to explore while the Q-table fills up then the agent exploits more than it explores in order to increase its reward. The learning rate should satisfy the following conditions $\sum_{t=0}^{\infty} \alpha_t = \infty$ and $\sum_{t=0}^{\infty} \alpha_t^2 < \infty$, e.g. $\alpha_t = 1/(t+1)$. Finally, note that $\alpha$ can also be taken as a constant less than one, and hence not satisfying the conditions above. However in practice, learning tasks occur over a finite time horizon hence the conditions are satisfied.
\paragraph{Q-learning.}
Q-learning has first been introduced by Watkins in 1989 \cite{Watkins1989}. In this algorithm, the target is equal to
\begin{equation} \label{eq:qtarget}
    T_{t+1} = R(t+1) + \gamma \max_{a'\in\set{A}} q_t \left(S(t+1) , a' \right) ,
\end{equation}
which is nothing but the algorithmic form of the Bellman equation in \eqref{eq:bellmanoptiq}. When the algorithm has converged, $T_{t+1}$ should be equal to $q_t(S(t),A(t))$, nullifying the difference term in \eqref{eq:qlearning}. Let us more closely examine how the Q-learning algorithm works.

The Q-table is a table with the states along the rows and columns representing the different actions we can take. At time step $t=0$, the Q-table is initialized to 0, i.e. $q_0(s(0),a(0)) = 0$, $\forall (s,a)\in \set{S}\times \set{A}$. A starting state is chosen randomly with a certain probability, i.e. $\prob{S(0)=s}{}$. At the beginning, the agent does not know the reward that each action will provide, hence it chooses one action randomly. This action leads to a new state $s'$ and a reward $r$ in a stochastic manner according to $p(r,s'\left|\right. s,a)$ if the problem is stochastic or according to deterministic rules otherwise. The immediate reward the agent receives by taking the action $a(t)$ at time $t$ is the variable $R(t+1)$ in \eqref{eq:qtarget}. Then the algorithm looks at the line represented by the state $S(t+1)=s'$ in the Q-table and chooses the value that is maximum in the line, i.e. this corresponds to the term $\max_{a'\in\set{A}} q_t(S(t+1),a')$ in \eqref{eq:qtarget}; among all possible actions from the next state we end up at $t+1$, i.e. $S(t+1)$, one chooses the one that leads to the maximum expected return. The difference in \eqref{eq:qlearning} acts as a gradient that allows to reinforce some actions in a given state or on the contrary, to dissuade the agent to take some other actions being in a given state. By choosing the learning rate, $\alpha_t = 1/(t+1)$ for instance, the agent will pay less and less attention over time to future expected returns in the update of the current estimation of the Q-table. If $\alpha=0$, the agent does not learn anymore while $\alpha=1$ means that the agent keeps an active learning behavior. Moreover, $\gamma$ in \eqref{eq:qlearning} is reminded to be the discounting factor and it defines how much the agent cares about future rewards, i.e. the ones that will be obtained starting from $S(t+1)$, compared to the immediate reward, i.e. $R(t+1)$.

Q-learning is an algorithm that explores and exploits at the same time. But which policy should be followed when updating the Q-table, i.e. how are the actions chosen at each time step? Actually, and this is the strength of Q-learning, it does not (so much) matter for the algorithm convergence. The $\epsilon-$greedy policy is, however, a widely used technique. It consists in randomly choosing an action with probability $\epsilon$ and the action that maximizes the current action-state value at time $t$ with probability $1-\epsilon$. Of course, $\epsilon$ can be kept constant or may vary during the learning in order to explore more at the beginning, i.e. $\epsilon\approx 1$, and exploit more after a while, i.e. $\epsilon \ll 1$. However, it has been shown that Q-learning makes the Q-table converge to $q_*$, and hence to the optimal policy, as soon as every action-state pair has been visited an infinite number of times, irrespective to the policy followed during the training \cite{Szepesvari1997,Even-Dar2003}. This property makes Q-learning an \emph{off-policy} procedure.

In some practical problems, MDP may present some terminal states, i.e. absorbing states, and the learning is done over several episodes. Once the agent reaches the terminal state the episode ends and the algorithm restarts at a random state and keep going to estimate the Q-table.

\paragraph{SARSA algorithm.}

SARSA is an algorithm that has been proposed in \cite{Rummery1994} and differs from Q-learning simply by the target definition in \eqref{eq:qlearning}. In SARSA, we use
\begin{equation} \label{eq:sarsatarget}
    T_{t+1} = R(t+1) + \gamma q_t \left(S(t+1) , A(t+1) \right).
\end{equation}
The difference with the Q-learning approach is that the next action $A(t+1)$ to be taken when the agent observes the next state $S(t+1)$ starting from the state $S(t)$ and having taken the action $A(t)$, is no longer the one that maximizes the next expected return from the next state $S(t+1)$. The action $A(t+1)$ has to be taken according to a policy, this is why SARSA is called an \emph{on-policy} method. The policy the agent follows when being in state $S(t)$, hence the choice of the action at time $t$, i.e. $A(t)$, is the \emph{behavior policy} while the choice of the action to be taken when being in state $S(t+1)$, i.e. $A(t+1)$, characterizes the \emph{target policy}. In Q-learning, the target policy was greedy w.r.t. $q_t$, i.e. the action that maximizes the next expected action-state value is chosen. In SARSA on the other hand, the agent updates the Q-table from the quintuple $\left( S(t) , A(t) , R(t+1) , S(t+1) , A(t+1) \right)$ where both behavior and target policies are $\epsilon-$greedy, i.e. the next action to be taken while observing state $S(t+1)$ is chosen randomly with the probability $\epsilon$ and the action maximizing $q_t(S(t+1),a')$ is taken with probability $1-\epsilon$. Under the assumption that the actions taken under the target policy are, at least occasionally, also taken under the behavior policy, SARSA converges to the optimal policy, i.e. the greedy policy w.r.t. $q_*$.

\subsection{Deep RL} \label{sec:deeprl}

The previous $Q$-learning or SARSA approaches are suitable when the dimensions of the state and action spaces of the problem are small or moderate. In that case, a look-up table can be used to update the $Q$ values. However, when the number of states or possible actions becomes large, the complexity and the storage needs of $Q$-learning become prohibitive. Instead of using a table for updating the action-state value function, one may search for approximating the action-state values with a suitable function $q_{\boldsymbol{\theta}} : \set{S}\times \set{A}\rightarrow \mathbb{R}$ with the vector parameter $\boldsymbol{\theta}$. The simplest approximation function is the linear one. This consists of finding a suitable mapping, $\psi : \set{S}\times \set{A}\rightarrow \mathbb{R}^d$, used to represent the \emph{features} of the action-state couple, where $d$ is the dimension of this feature representation \cite{Szepesvari2010}. The entries of the vector $\psi(s,a)$ are the \emph{basis functions} and they span the space in which the Q function is approached. The linear approximation of the action-value function is hence $q_{\boldsymbol{\theta}}(s,a) = \boldsymbol{\theta}^T \boldsymbol{\psi}(s,a)$ and will be used to approximate the optimal action-state value.

If an oracle would give us the action-state function under the policy $\pi$, one could compute an error function between a target and a prediction as follows:
\begin{equation} \label{eq:lossoracle}
    L(\boldsymbol{\theta}) = \Exp{\left(q_{\pi}(S,A) - \boldsymbol{\theta}^T \boldsymbol{\psi}(S,A)\right)^2}{\pi} ,
\end{equation}
where the expectation is taken over the joint distribution of the state, reward and action. The weights $\boldsymbol{\theta}$ can be updated using the gradient of the loss function such as
\begin{equation} \label{eq:sgdoracle}
    \boldsymbol{\theta}_{t+1} = \boldsymbol{\theta}_t + \alpha_t \Exp{\left(q_{\pi}(S,A) - \boldsymbol{\theta}^T \boldsymbol{\psi}(S,A)\right) \boldsymbol{\psi}(S,A)}{\pi} .
\end{equation}
However, the agent never knows in general the true value of the objective, i.e. the true action-state value function under the policy $\pi$. It can only estimate this value. For the Q-learning algorithm, $q_{\pi}(S,A)$ in \eqref{eq:sgdoracle} is substituted by expression in \eqref{eq:qtarget}, where the estimation of the action-state function is replaced by its linear approximation such that
\begin{equation} \label{eq:sgdq}
    \boldsymbol{\theta}_{t+1} = \boldsymbol{\theta}_t + \alpha_t \left(R(t+1) + \gamma \max_{a'\in \set{A}} \boldsymbol{\theta}^T \boldsymbol{\psi}(S(t+1),a') - \boldsymbol{\theta}^T \boldsymbol{\psi}(S,A)\right) \boldsymbol{\psi}(S,A) .
\end{equation}
However, the features extraction phase, i.e. constructing the function $\boldsymbol{\psi}$, can be very complex if the problem dimension is very large \cite{Szepesvari2010}.


One may observe that when passing from \eqref{eq:sgdoracle} to \eqref{eq:sgdq}, we did not limit ourselves to substitute $q_{\pi}(S,A)$ by $R(t+1) + \gamma \max_{a'\in \set{A}} \boldsymbol{\theta}^T \boldsymbol{\psi}(S(t+1),a')$ but the expectation operator also vanished. There are only random samples from a given database (a batch) to compute the sequence $\left\{\boldsymbol{\theta}\right\}_t$. The latter is a random sequence and hence we are not sure that this will decrease the value of the loss at each step. However, one can show that this will decrease the loss in \emph{average}. This is the principle of the stochastic gradient descent \cite{Bottou1999}.

\paragraph{Deep Q-learning.} The approximation function can be non-linear, but has to be differentiable w.r.t. the parameters of the approximation function. The principle of the deep Q-learning is simple and consists of designing a neural network that outputs all the action-state values for all possible actions in a given state $s$, $q_{\boldsymbol{\theta}}(s,\cdot)$, $\forall s\in \set{S}$. In other words, if the state space is of dimension $n$, i.e. each state is represented by a vector of $n$ components, and if there are $m$ possible actions for each state, the neural network is a mapping from $\mathbb{R}^n$ to $\mathbb{R}^m$. The idea to use a deep neural network to approximate the Q-function in a reinforcement learning setting has been first introduced by Mnih \emph{et al.} in \cite{Mnih2015}, where authors proposed to use a deep convolution neural network to learn to play Atari games.

We will not discuss here in details the different neural networks and deep learning in general and the interested reader may refer to some reference books dealing with neural networks and the basics of deep learning, such as the book of Courville, Goodfellow and Bengio \cite{Goodfellow2016}. An ANN is made of (artificial) neurons that are linked through several layers. There exists several kind of neural networks, such as the feedforward neural network (FNN), the recurrent neural network, convolutional neural network as mentioned above and many others (cf. for instance \cite{Chen2019} for a classification of ANN). In the following, we will just refer to FNN.
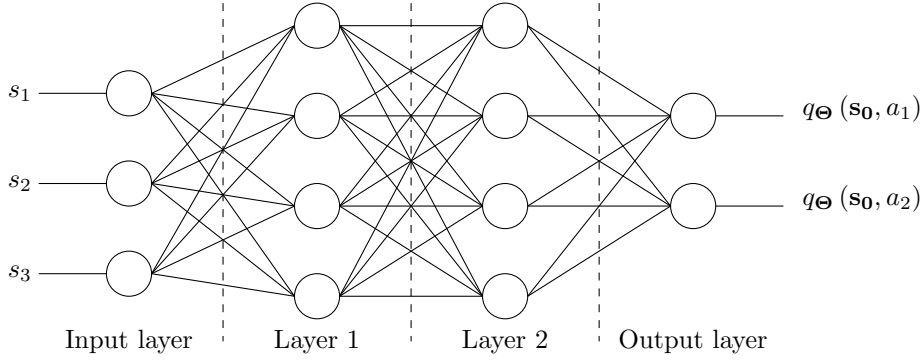
\begin{figure}[htbp]
\begin{center}
    \begin{tikzpicture}
        \node at(-3.75 , -2){Input layer} ;
        \draw (-4.95 , 1.3) -- ( -4.05 , 1.3) ;
        \node at(-5.2 , 1.3){$s_1$};
        \draw (-4.95 , 0.1) -- ( -4.05 , 0.1) ;
        \node at(-5.2 , 0.1){$s_2$};
        \draw (-4.95 , -1.1) -- ( -4.05 , -1.1) ;
        \node at(-5.2 , -1.1){$s_3$};
        \draw (-3.75 , 1.3) circle (0.3) ;
        \draw (-3.75 , 0.1) circle (0.3) ;
        \draw (-3.75 , -1.1) circle (0.3) ;
        \node at(-1.25 , -2){Layer 1} ;
        \draw (-1.25 , 2.2) circle (0.3) ;
        \draw (-1.25 , 1) circle (0.3) ;
        \draw (-1.25 , -0.2) circle (0.3) ;
        \draw (-1.25 , -1.4) circle (0.3) ;
        \node at(1.25 , -2){Layer 2} ;
        \draw (1.25 , 2.2) circle (0.3) ;
        \draw (1.25 , 1) circle (0.3) ;
        \draw (1.25 , -0.2) circle (0.3) ;
        \draw (1.25 , -1.4) circle (0.3) ;
        \node at(3.75 , -2){Output   layer} ;
        \draw (3.75 , 1) circle (0.3) ;
        \draw (3.75 , -0.2) circle (0.3) ;
        \draw (-3.45 , 1.3) -- (-1.55 , 2.2);
        \draw (-3.45 , 1.3) -- (-1.55 , 1);
        \draw (-3.45 , 1.3) -- (-1.55 , -0.2);
        \draw (-3.45 , 1.3) -- (-1.55 , -1.4);
        \draw (-3.45 , -1.1) -- (-1.55 , 2.2);
        \draw (-3.45 , -1.1) -- (-1.55 , 1);
        \draw (-3.45 , -1.1) -- (-1.55 , -0.2);
        \draw (-3.45 , -1.1) -- (-1.55 , -1.4);
        \draw (-3.45 , 0.1) -- (-1.55 , 2.2);
        \draw (-3.45 , 0.1) -- (-1.55 , 1);
        \draw (-3.45 , 0.1) -- (-1.55 , -0.2);
        \draw (-3.45 , 0.1) -- (-1.55 , -1.4);
        \draw (-0.95 , 2.2) -- (0.95 , 2.2);
        \draw (-0.95 , 2.2) -- (0.95 , 1);
        \draw (-0.95 , 2.2) -- (0.95 , -0.2);
        \draw (-0.95 , 2.2) -- (0.95 , -1.4);
        \draw (-0.95 , 1) -- (0.95 , 2.2);
        \draw (-0.95 , 1) -- (0.95 , 1);
        \draw (-0.95 , 1) -- (0.95 , -0.2);
        \draw (-0.95 , 1) -- (0.95 , -1.4);
        \draw (-0.95 , -0.2) -- (0.95 , 2.2);
        \draw (-0.95 , -0.2) -- (0.95 , 1);
        \draw (-0.95 , -0.2) -- (0.95 , -0.2);
        \draw (-0.95 , -0.2) -- (0.95 , -1.4);
        \draw (-0.95 , -1.4) -- (0.95 , 2.2);
        \draw (-0.95 , -1.4) -- (0.95 , 1);
        \draw (-0.95 , -1.4) -- (0.95 , -0.2);
        \draw (-0.95 , -1.4) -- (0.95 , -1.4);
        \draw (1.55 , 2.2) -- (3.45 , 1);
        \draw (1.55 , 2.2) -- (3.45 , -0.2);
        \draw (1.55 , 1) -- (3.45 , 1);
        \draw (1.55 , 1) -- (3.45 , -0.2);
        \draw (1.55 , -0.2) -- (3.45 , 1);
        \draw (1.55 , -0.2) -- (3.45 , -0.2);
        \draw (1.55 , -1.4) -- (3.45 , 1);
        \draw (1.55 , -1.4) -- (3.45 , -0.2);
        \draw (4.05 , 1) -- (4.95 , 1) ;
        \node at(6 , 1.1){$q_{\boldsymbol{\Theta}}\left(\mathbf{s_0},a_1\right)$} ;
        \draw (4.05 , -0.2) -- (4.95 , -0.2) ;
        \node at(6 , -0.1){$q_{\boldsymbol{\Theta}}\left(\mathbf{s_0},a_2\right)$} ;
        \draw[dashed] (-2.5 , 2.5) -- (-2.5 , -2) ;
        \draw[dashed] (0 , 2.5) -- (0 , -2) ;
        \draw[dashed] (2.5 , 2.5) -- (2.5 , -2) ;
    \end{tikzpicture}
\end{center}
\caption{Feedforward neural network} \label{fig:fnn}
\end{figure}
The main ingredients of an FNN in our case are an input layer that accepts the states of the decision process, or more specifically, a \emph{representation} of the states, an output layer that returns the estimated action-state values for all actions for a given state at the input and several hidden layers between both as illustrated on Fig. \ref{fig:fnn} where two hidden layers has been considered. The input layer is made of $n_0$ neurons, and takes the samples $\boldsymbol{s}_0 \in \mathbb{R}^n$ as an entry. The layer $\ell = 1, \cdots, L$ has $n_{\ell}$ neurons. The $L+1$ layer is the output layer, and has $n_{L+1}$ neurons. It outputs the approximation of the Q-function, for the entry $\boldsymbol{s}_0$, i.e. $q_{\boldsymbol{\Theta}}(\boldsymbol{s}_0,a_i)$ of each action $a_i$, $i=1,\cdots, n_{L+1}$. $\boldsymbol{\Theta}$ represents the parameters of the neural network that will be explained hereafter.

For all $\ell = 1, \cdots, L+1$, the vector output at the layer $\ell$, $\boldsymbol{s}_{\ell} \in \mathbb{R}^{n_{\ell}}$, is
\begin{equation} \label{eq:neuout}
    \boldsymbol{s}_{\ell} = f_{\ell} \left(\boldsymbol{\theta}_{\ell} \boldsymbol{s}_{\ell-1} + \boldsymbol{b}_{\ell}  \right)
\end{equation}
where $\boldsymbol{\theta}_{\ell}\in \mathbb{R}^{n_{\ell-1}\times n_{\ell}}$ is the matrix of weights between the neurons of layer $\ell-1$ and the layer $\ell$, i.e. $\theta^{\ell}_{i,j}$ is the weight between the $i-$th neuron of layer $\ell-1$ and the $j-$th neuron of layer $\ell$, $\boldsymbol{s}_{\ell-1}$ is the signal output of the layer $\ell-1$. Moreover, $f_{\ell}$ is the \emph{activation function} at the layer $\ell$ whose objective is to perform a non-linear operation on the linear combination of the signal output of the previous layer. The function applies at each neuron of layer $\ell$, i.e. the function applies to each entry of the vector in the argument of $f_{\ell}$ in \eqref{eq:neuout}. The activation function aims at enabling the signal output. It exists several kind of activation functions, e.g. sigmoidal, hyperbolic tangent, ReLU (Rectified Linear Unit) and ELU (exponential linear unit), the reader may refer to \cite{Zappone2019} for a more detailed description. Finally, $\boldsymbol{b}_{\ell}$ is the bias term of the neurons of the layer $\ell$ that can change the threshold of the input signal value at which the neurons enable the output signal. The parameters of the FNN is made of the succession of matrix weights and bias terms between each layers and is denoted: $\boldsymbol{\Theta} = \left\{ \boldsymbol{\theta}_{\ell} , \boldsymbol{b}_{\ell} \right\}_{\ell=1}^L$.

In order to make a neural network learn, a loss function should be defined between a target, i.e. a desired output, and the actual output obtained at iteration $t$. It can be defined as in \eqref{eq:lossoracle} by using the target in Q-learning, i.e. \eqref{eq:qtarget}:
\begin{equation} \label{eq:drlloss}
    L_t(\boldsymbol{\Theta}_t) = \Exp{\left( R(t+1) + \gamma \max_{a'} q_{\boldsymbol{\Theta}_t^{-}}\left(S(t+1),a'\right) - q_{\boldsymbol{\Theta}_t}\left(S(t),A(t)\right) \right)^2}{S,A,R,S'} ,
\end{equation}
where $\boldsymbol{\Theta}_t$ is the set of parameters at iteration $t$ and $\boldsymbol{\Theta}_t^{-}$ is the network parameters used to compute the target at iteration $t$. Deriving the loss function w.r.t. the network parameters we obtained a generalized stochastic gradient descent for non linear Q-function approximation as
\begin{multline} \label{eq:sgddeep}
    \boldsymbol{\Theta}_{t+1} = \boldsymbol{\Theta}_t + \alpha_t \left(R(t+1) + \gamma \max_{a'\in \set{A}} q_{\boldsymbol{\Theta}^{-}_t}(S(t+1),a') - q_{\boldsymbol{\Theta}_t}(S(t),A(t))\right) \times \\ \nabla_{\boldsymbol{\Theta}_t} q_{\boldsymbol{\Theta}_t}(S(t),A(t)) ,
\end{multline}
which decreases the expected loss under certain conditions. The gradient in \eqref{eq:sgddeep} should be understood as the gradient w.r.t. the weights keeping the bias constant and also the gradient w.r.t. the bias keeping the weights constant.

However, this method may diverge if naively implemented. Indeed, if the samples $\left(s,a,r,s'\right)$ are obtained by sampling the Markov chain at each successive transition, the data given to the neural network will be correlated and non i.i.d and will be not ideal for training the neural network. A solution is to store the experience tuples $\left(s(t),a(t),r(t+1),s(t+1)\right)$ into a memory pooled over many episodes (an episode ends when a terminal state is reached) and to uniformly choose at each time instant a tuple from the memory in order to update \eqref{eq:sgddeep}. This is what is called \emph{experience replay} and it breaks correlation among the samples \cite{Mnih2015}. Moreover, if the target network $\boldsymbol{\Theta}^{-}$, used to retrieve the Q-values, is updated after each iteration with the new network computed at time $t$, then the policy may oscillate and data values may switch from an extreme to another and the parameters could diverge. The idea is to freeze the target network $\boldsymbol{\Theta}^{-}$ during a certain number of time steps $T$ and to update the target network with $\boldsymbol{\Theta}$ in \eqref{eq:sgddeep} after $T$ steps. This allows for reducing the oscillations or the risk of divergence.

There is another unwanted effect of the Q-learning that is the overestimation of the action-state values. This bias leads the algorithm to perform poorly in some stochastic environments and comes from the fact that the $\max$ operator is used to estimate the maximum expected value. Indeed, the $\max$ operator in \eqref{eq:qtarget} aims at estimating the value in \eqref{eq:actionstatevalue} for the next state $S(t+1)$ which is an expectation. This method, often called the \emph{single estimator}, has a positive bias that can be shown to follow from Jensen's inequality. Van Hasselt proposed in \cite{Hasselt2010} to use a \emph{double estimator} technique to unbiase the estimation of the maximum expected value that occurs in the Bellman's equation of the action-state value function. The idea is to create two sets of unbiased estimators w.r.t. the expectation, i.e. $q^A$ and $q^B$, that will be applied on two sets of samples, i.e. $\set{A}$ and $\set{B}$, such that the estimators are unbiased w.r.t. the mean on these samples. $\mathcal{A}$ and $\mathcal{B}$ contain the samples associated to the random variables $q^A(S(t),\cdot)$ and $q^B(S(t),\cdot)$ respectively. The maximum expected value of the first estimator, i.e. $q^A$ is estimated with the $\max$ operator on the set of the experiences $\set{A}$, i.e. $\max_a q^A(s',a) = q^A(s',a^*)$ as in the regular Q-learning algorithm. Then, we use the action $a^*$ on the estimator of the Q-function on the set $\set{B}$, as an estimation of the maximum expected value of $\max_a \Exp{q^A(S',a)}{}$. A similar update is performed with $b^*$ on $q^{B}$ and using $q^A$. The iterative system of equations in the double Q-learning are such as \cite[Algorithm 1]{Hasselt2010}
\begin{eqnarray}
    q_t^A(S(t),A(t)) & \leftarrow & q^A_t(S(t),A(t)) + \alpha_t\left[ R(t+1) + \gamma q^B_t(S(t+1),a^*) - q^A_t(S(t),A(t)) \right] , \nonumber \\
    q_t^B(S(t),A(t)) & \leftarrow & q^B_t(S(t),A(t)) + \alpha_t\left[ R(t+1) + \gamma q^A_t(S(t+1),b^*) - q^B_t(S(t),A(t)) \right] . \nonumber
\end{eqnarray}
The principle of the \emph{double Q-learning} can be applied to any approximation techniques of the action-state function, and in particular when a deep neural network is employed \cite{Hasselt2016}.

\subsection{Multi-armed bandits} \label{sec:mab}

\subsubsection{Fundamentals}

A multi-armed bandit (MAB) model holds its name from a slot machine with several levers\footnote{Or equivalently several one-armed gambling machines}, that the player/user activates in the hope to hit the prize. Each machine has a certain probability to deliver the money to the player. The goal of the agent, to keep the terminology used in the RL framework, is to play the machine that gives the maximum expected gain in the long run. Mathematically, a MAB model is a collection of $K$ random variables $R_i$, $i=1,\cdots, K$, where $i$ denotes the "arm" of the bandit, each distributed as $P_{R_i}$, unknown to the agent. The player sequentially chooses an arm, i.e. the action of the agent $\left\{A(t)\right\}_{t \geq 0}$ and collects rewards over time $\left\{R(t+1)\right\}_{t\geq 0}$. When the agent pulls arm $a$ as its action at time $t$, i.e. $A(t) = a$, it gets a reward randomly drawn from the distribution $P_{R_a}$, i.e. $R(t) \sim P_{R_a}$. The goal of the agent is to maximize the expected rewards obtained up the time horizon $T$, i.e. $\Exp{\sum_{t=1}^T R(t)}{P_{R}}$. By denoting $\mu_i$ the expectation of arm $i$, i.e. $\mu_i = \Exp{R_i}{}$, the agent should play as much as possible the arm with the maximum mean reward, $\mu^* = \arg \max_i \mu_i$, and less as possible the suboptimal arms. However, the arm with the maximum mean reward is of course unknown to the agent, in general, and the agent has to make decisions, defining its policy, based only on the past observations. To do so, the agent has to explore sufficiently in order to accumulate information on the rewards given by the arms, and also has to exploit the best arms, i.e. those that have given the highest cumulated rewards so far. This is the famous exploration-exploitation tradeoff we already mentioned above that each RL algorithm has to deal with.

One could consider the MAB as a special case of MDP described in Subsection \ref{subsec:mdp} where only a single state is considered. The conditional probability distribution defining the dynamic of this particular MDP in \eqref{eq:srdist} reduces to
\begin{equation}
    P_{R(t+1),S(t+1)\left|\right. S(t),A(t)}(r,s'\left|\right. s,a) = P_{R(t)\left|\right.A(t)}(r\left|\right. a) \defeq P_{R_a}.
\end{equation}
In the definition above, the state transition vanishes because there is only one state in the MDP and the reward is immediately obtained after pulling the lever $a$.

There are two schools of thought in MAB, or two approaches: the Bayesian, proposed by Thompson \cite{Thompson1933}, and the frequentist approach e.g. \cite{Robbins_1952,Lai_1985}. When the distribution of arms depends on a parameter $\vec{\theta}$, the MAB framework is said to be \emph{parametric}, i.e. the distribution of the MAB is $P_{\boldsymbol{R_\theta}} = \left(P_{R_{\theta_1}}, \cdots, P_{R_{\theta_K}}\right)$ and $\boldsymbol{\theta} = \left(\theta_1,\cdots,\theta_K\right)$ is the vector parameter of the MAB. In the Bayesian setting, $\boldsymbol{\theta}$ is a random variable drawn from a prior distribution $P_{\boldsymbol{\theta}}$. In i.i.d. scenario, e.g. the rewards are drawn from a Bernouilli distribution\footnote{We will see later the Markovian setting}, and conditionally to $\theta_a$, $R_a(t)\sim P_{R_a}$ with mean $\mu_a$ and the elements of the series $\left\{R_a(t)\right\}_{t, a}$ are independent. In the frequentist approach, $R_a(t)$ has the same properties as in the Bayesian setting, but where $\boldsymbol{\theta}$ is no longer random. Instead it is an unknown deterministic parameter. We will not discuss the difference between Bayesian and frequentist approaches further and the interested reader may consult the excellent treatise in \cite{Kaufmann2014} for more details about the both approaches. In the following, we will focus on the frequentist approach for which the notion of \emph{regret} is defined \cite{Auer_2002}.

\paragraph{Regret.} The regret under the time horizon $T$ can be understood rather intuitively: this is the difference between the average expected reward one would obtain if one always plays the optimal arm and the average expected reward obtained following a policy $\pi$, which is the sequential series of actions $A(0), A(1), \cdots A(T-1)$, different from the optimal one. By denoting the index of the optimal arm as $o = \arg \max_{i} \mu_i$ and its expected reward $\mu^* = \max_{i} \mu_i \defeq \Exp{R_o(t)}{}$, the regret may be written as
\begin{equation} \label{eq:regret1}
    \mathcal{D}_{\boldsymbol{\mu}}(T) = T\mu^* - \Exp{\sum_{t=1}^T R(t)}{P_R}.
\end{equation}
Note that the regret depends on the parameter $\boldsymbol{\mu} = \left(\mu_1,\cdots, \mu_K \right)$. The regret can also be seen in a different, but totally equivalent, way. Let us consider several experiments of duration $T$, each arm $i=1,\cdots, K$ will be played $N_i(T)$, which is a random variable for all $i$, with the expected value $\Exp{N_i(T)}{}$. For instance let us assume that $K=3$ and $T=9$. During the first experiment, arm 1 has been played 4 times, arm 2 has been played 3 times and arm 3, 2 times\footnote{The sequence of the arm selection depends on the policy the agent follows.}. Assuming a stationary setting, i.e. the distribution of the regret does not vary in time, the regret in \eqref{eq:regret1} becomes $T\mu^* - 4\Exp{R_1}{} - 3\Exp{R_2}{} -2\Exp{R_3}{}$ or $T\mu^* - 4\mu_1 - 3 \mu_2 - 2\mu_3$. The second experiment gives $(5,2,2)$ for arms 1, 2 and 3 respectively and so on. If $n$ experiments are run, the empirical average of the regret over the experiments, $\overline{D_{\mu}(T)}$, in our example gives
\begin{equation}
    \overline{D_{\mu}(T)} = T\mu^* - \left(\overline{N_1(T)}\mu_1 + \overline{N_2(T)}\mu_2  + \overline{N_3(T)}\mu_3\right) ,
\end{equation}
where $\overline{N_i(T)} = \frac{1}{n} \sum_{j=1}^n N_i(j ; T)$ is the empirical average of the number of times arm $i$ has been pulled at time $T$. When $n\rightarrow \infty$, $\overline{N_i(T)} \rightarrow \Exp{N_i(T)}{}$, and hence the regret in \eqref{eq:regret1} can be expressed as a function of the average number of times each arm has been played:
\begin{equation} \label{eq:regret2}
    \mathcal{D}_{\boldsymbol{\mu}}(T) = \sum_{i=1}^K \left(\mu^* - \mu_i\right) \Exp{N_i(T)}{} ,
\end{equation}
where the expectation depends on $\boldsymbol{\mu}$. From \eqref{eq:regret2}, it seems clear that a policy should play as much as possible the best arm, and as less as possible the suboptimal ones. Lai and Robbins have shown, \cite[Th. 2]{Lai_1985}, that any policy cannot have a better regret than a logarithmic one asymptotically, i.e. when $T\rightarrow \infty$. It is equivalent to say that the average number of plays of suboptimal arms is logarithmically bounded when $T\rightarrow \infty$, i.e.
for all $i$ such as $\mu_i<\mu^*$ and if $\boldsymbol{\mu} \in \left[0,1\right]^K$ we have
\begin{equation}
    \lim \inf_{T\rightarrow \infty} \frac{\Exp{N_i(T)}{}}{\log T} \geq \frac{1}{D\left( P_{R_i} \left| \right| P_{R_o}\right)}
\end{equation}
where $D\left(P_{R_i}\left|\right| P_{R_o}\right)$ is the Kullback-Leibler divergence between the reward distributions of arm $i$ and the optimal arm respectively. This result gives the bound of fundamental performance of any sequential policy in MAB settings, but it does not give any insight on how to explicitly design a policy that achieves this bound. Several algorithms have been proposed that are known to be order optimal, i.e. whose the regret behaves logarithmically with time. We will not detail all of them in this chapter but we will focus on a particular class that is applicable to physical layer communications: the upper confidence bound (UCB) algorithm.

\paragraph{Upper confidence bound algorithm.} The UCB algorithms are based on the computation of an index for each arm and selecting the arm with the highest index. The index is composed of two terms. The first is an estimate of the average reward of arm $i$ at time $t$, i.e. $\hat \mu_i(t) = \frac{1}{N_i(t)}\sum_{n=1}^{N_i(t)} R_i(n)$, and the second term is a measure of the uncertainty of this estimation. The UCB algorithm consists of choosing the maximum of this uncertainty to compute the index. Auer \emph{et al.} proposed several algorithms in \cite{Auer_2002} for bounded expected rewards in $\left[0,1\right]$. The most famous algorithm is UCB1 that allows one to choose arm $a(t+1)$ at the next time such that
\begin{equation}
    a(t+1) = \arg \max_i \left[\hat \mu_i(t) + \sqrt{\frac{2 \log t}{N_i(t)}} \right] ,
\end{equation}
where the second term in the square root can be obtained using Chernoff-Hoeffding inequality, which represents the upper bound of the confidence interval in the estimation of $\hat \mu_i$. The more arm $i$ is played, i.e. as $N_i(t)$ increases, the smaller the confidence interval, which means the index value relies on the average value of the cumulated reward obtained so far. When arm $i$ is played less the second term is larger, which encourages the agent to play another arm. The second term allows for exploration while the first term encourages the exploitation of the arm that has given the largest rewards so far. Moreover, thanks to the $\log$ term that is unbounded with time, all arms will be played asymptotically. Note that other UCB algorithms can perform very well in practice such as the Kullback-Leibler UCB (KL-UCB) for which a non-asymptotic regret bound can be proved \cite{cappe_2013}. Other algorithms have been proposed in order to deal with non-stationary environments in which the parameters of the distribution may change with time \cite{Oksanen2015}.

\subsection{Markovian MAB}

Historically, the first bandits studied were binary, i.e. the rewards are drawn from a Bernouilli distribution. Actually the result of Lai and Robbins in \cite{Lai_1985} is quite general and holds for $\boldsymbol{\mu} \in \left[0,1\right]^K$ for any distribution of the arms $P_{R_i}$. However, the independence of the rewards is an important assumption that does not necessarily hold in many practical problems.

In particular, the case of Markovian rewards is of practical interest in wireless communication. Each arm in Markovian MAB is characterized by a non-periodic Markov chain with finite state space $\set{S}_i$. For all arm $i$, the agent receives a positive reward $R_i(s)$ that depends on the state observed for arm $i$. The change from a state to another follows a Markovian process, under the conditional probability $P_{S_i(t+1)\left|\right. S_i(t)}(s'_i\left|\right. s_i) \defeq p(s'_i\left|\right. s_i)$. The stationary distribution of the Markov chain of arm $i$ is denoted as $\boldsymbol{P}_i = \left\{P_{S_i}(s) \defeq p_i(s), S_i \in \set{S}_i \right\}$. Each arm $i$ has an expected reward that is
\begin{equation}
    \mu_i = \sum_{s\in \set{S}_i } r_i(s) p_i(s) \defeq \Exp{R_i}{P_{S_i}}
\end{equation}
The goal of the agent is still to find a policy $\pi$, i.e. the sequential observation of arms, that minimizes the regret over the time, which is defined in \eqref{eq:regret1} or \eqref{eq:regret2}. When the agent observes an arm $i$ at $t$, it samples a Markovian process which evolves with time. A particular attention has to be paid on the status of Markov chains that are not observed that leads to the distinction between the \emph{rested} and \emph{restless} cases.

\paragraph{Rested MAB.} A Markovian MAB is qualified as rested, when only the Markov chain of the arm that is played evolves, the others remaining frozen. This assumption is strong since the Markov chains of the arms that are not observed do not evolve with time and it does not matter how much time elapsed between two consecutive visits to a given arm. The authors in \cite{anantharam_1987_m} were the first to be interested in the fundamental performance in terms of regret of Markovian MAB with multiple plays\footnote{When the player may pull more than one arm or if multiple players are considered (without collisions).} and they proposed a policy that is \emph{asymptotically} efficient, i.e. that achieves the regret lower bound asymptotically. The authors in \cite{tekin_2012} showed that a slightly modified UCB1 achieves a logarithmic regret uniformly over time in this setting.

\paragraph{Restless MAB.} A Markovian MAB is considered as restless, if the Markov chains of all arms evolve with time, irrespective of which arm is played. This assumption implies radical changes in the regret analysis because the state we will observe when pulling an arm $i$ at $t+1$ directly depends on the time elapsed since the last visit to this arm. This is because the reward distribution we get by playing an arm depends on the time elapsed between two consecutive plays of this arm and since arms are not played continuously, the sample path experienced by the agent does not correspond to a sample path followed when observing a discrete time homogeneous Markov chain. The solution came from \cite{tekin_2011} where the authors proposed a \emph{regenerative cycle algorithm} to deal with the discontinuous observation of evolving Markov chains. In practice, the agent still keeps going to apply the UCB1 algorithm, introduced earlier, but computes the index only on the samples the agent has collected when observing a given arm $i$. This structure requires one to observe an arm during a certain amount of time before computing the UCB1 index. The interested reader may refer to \cite{tekin_2011} for further details and \cite{tekin_2012} for the extension to multiple plays. This setting is particularly interesting because it finds a natural application in wireless communications with the opportunistic spectrum access scenario when the state of the bands the user has to access evolves independently of the action the user takes. For example, the band may be sporadically occupied by another system or the propagation condition may evolve with time.

\subsubsection{Contextual MAB}

Contextual MAB generalizes the classical concept introduced above towards more general reinforcement learning. Conventional MAB does not take advantage of any knowledge about the environment. The basic idea of contextual MAB is to condition the decision making on the state of the environment. This allows for making the decisions based both on the particular scenario we are in and the previous observations we have acquired. A contextual MAB algorithm observes a context in the form of useful side information, followed by a decision by choosing one action from the set of alternative ones. It then observes an outcome of that decision which defines the obtained reward. In order to benefit from the context information, there needs to exist dependency between the expected reward of an action and its context. The goal of learning is to maximize a cumulative reward function over the time span of interest. The side information in physical layer communications may be a feature vector containing for example location and device information, experienced interference, received signal strength fingerprint, channel state information (CSI) of a particular user, or potential priority information. Such side information would allow for selection of a base station, service set or antennas such that higher rewards are achieved. Extensions of well-known UCB algorithms have been developed for contextual MAB problems. Since the context plays an important role in the dynamic of contextual MAB, it may be crucial to detect the change of a context in order to adapt the policy to this change. Methods based on statistical multiple change-point detection and sequential multiple hypothesis testing may be used for that purpose \cite{Poor2008,Nitzan2020}.

\subsubsection{Adversarial MAB}

Adversarial bandit problems are defined using sequential game formulation. The adversarial model means that the decisions may lead to the worst possible payoff instead of optimistic view of always making choices that lead to optimal payoff.  The problems are modeled assuming deterministic and uninformed adversary and that the payoffs or costs for all arms and all time steps of playing the arms are chosen in advance. The adversary is often assumed to be uninformed so that it makes its choices independent of the previous outcomes of  the strategy. At each iteration, a MAB agent chooses an arm it plays while an adversary chooses the payoff structure for each arm. In other words, the reward of each arm is no longer chosen to be stochastic but they are \emph{deterministically assigned} to an \emph{unknown} sequence, $\mathbf{r}(1), \mathbf{r}(2), \cdots$ where $\mathbf{r}(t) = \left(r_1(t), \cdots, r_K(t)\right)^T$ and $r_i(t) \in \set{R} \subset \mathbb{R}$ is the reward of the $i-$th arm at time $t$. By denoting the policy $\pi$ that maps a time slot to an arm index to play at the next time slot\footnote{Actually the policy is a mapping from the set of indices and rewards obtained so far to the next index, i.e. $\pi: \left( \left\{1,\cdots,K\right\}\times \set{R} \right)^{t-1} \rightarrow \left\{1,\cdots,K\right\}$}, $\pi(1), \pi(2), \cdots$ is the sequence of plays of the agent. Considering a time-horizon $h$, the goal of the agent is to minimize the regret
 \begin{equation} \label{eq:advregret}
 \mathcal{D}_h = \max_{j\in \left\{1, \cdots, K\right\}} \sum_{t=1}^{h} r_j(t) - \Exp{\sum_{t=1}^{h} R_{\pi(t)}(t)}{\pi} ,
 \end{equation}
where the adversary chooses $r_1(1), r_2(1)...,r_k(t)$ and the player strategy chooses actions $\pi(t)$. Since the policy is commonly assumed to be random, the regret is also random (so the notation $R_{\pi(t)}(t)$), and the expectation is taken over the distribution of the policy. Adversarial bandits is an important generalization of the bandit problem since no assumptions on the underlying distributions are made, hence the name adversarial. The player has access to the trace of rewards for the actions that the algorithm chose in previous rounds, but does not know the rewards of actions that were not selected. Widely used methods such as UCB are not suitable for this problem formulation. The exponential-weight algorithm called {\em Exp3} for exploration and exploitation is widely used for solving Adversarial bandits problems \cite{Auer_2003, Stoltz2005}.
\section{RL at PHY layer} \label{sec:apps}

In this section, the concepts of Q-learning, deep-learning and MAB are illustrated through some examples deriving from communication problems addressed in literature. The first example in Section \ref{sec:qlearex} deals with power management under queuing delay constraint in a point-to-point wireless fading communication channel and derives from \cite{Ngo2010,Mastronarde2013}. This problem has been widely studied in literature under various formalization including with MDP (cf. the discussion in Section \ref{sec:relatedworks}) and a rather simple toy example can be extracted from it. In Section \ref{sec:drlex}, we present two examples to illustrate the reinforcement learning in large dimension, i.e. optimal caching in a single cell network in Section \ref{sec:linearrl} and the extension of the single user power-delay management problem dealt with in Section \ref{sec:qlearex} to the multi-user case with large state and action spaces in Section \ref{sec:dqnrl}. Finally Section \ref{sec:mabex} illustrates the use of MAB with the opportunistic spectrum access (OSA) problem and green networking issue in Sections \ref{sec:mabosa} and \ref{sec:mabgreennet}, respectively. This section ends with experimental results and proof-of-concept to validate the MAB principle applied to the OSA issue. These examples do not aim at providing the full solution of the problem raised, which can be found in the related literature, but rather a simple problem statement with explicit action and state spaces and the reward function that can be chosen to solve the problem with a reinforcement learning strategy.

\subsection{Example with Q-learning} \label{sec:qlearex}

Let us consider a point to point communication system over a block fading channel, i.e. the channel gain is assumed to be constant on a slot of duration $\Delta t$ and changes from one slot to another according to the distribution $P_{H(t+1)\left|\right.H(t)}(h'\left|\right. h)$ with $H\in \set{H}$ where $\set{H}$ is assumed to be a finite countable set. At each time slot, the transmitter can send a packet or remain silent and it can also choose its transmit power as well as its modulation and coding scheme, e.g. a 16-QAM with a convolutional code of rate 1/2. At each time slot, a certain number of bits is generated by the application layer and stored in a buffer waiting for their transmission. The transmitter aims at sending the highest number of bits as possible with minimal power consumption while limiting the waiting time in the buffer.

At each time slot $t$, $N(t)$ new bits are generated and stored in the buffer before being transmitted. $\left\{N(t)\right\}_{t\in\mathbb{N}}$ are i.i.d. random variables with the distribution $P_N(n)$. The buffer state $B(t)\in \set{B}=\left\{0,1,\cdots,B_{\max}\right\}$ represents the number of bits stored in the queue at time $t$ and $B_{\max}$ is the maximal buffer size. At each time slot, transmitter chooses $\beta(t)\in \left\{1,\cdots, B(t)\right\}$ bits\footnote{Only a maximum of $B(t)$ bits can be encoded and sent at time $t$.} to be transmitted and encodes them into a codeword of length $n_c$ channel uses, assumed to be fixed. The rough spectral efficiency of the transmission is hence $\rho(t) = \beta(t)/n_c$. Moreover, transmitter chooses its power level $P_{\text{tx}}(t)\in \left\{0,p_{1},\cdots, p_{\max}\right\}$ and hence the total power consumed at $t$ is
\begin{equation}
P_{\text{tot}}(t) = p_{\text{on}} + \alpha^{-1} P_{\text{tx}}(t) ,
\end{equation}
where $p_{\text{on}}$ is the static power consumed by the electronic circuits and $\alpha\in \left]0,1\right]$ the efficiency of the power amplifier. One can define the state space of the power consumption as $\set{P} = \left\{p_{\text{on}},\cdots,p_{\text{on}} + \alpha^{-1} p_{\max}\right\}$. 

The codeword error probability, a.k.a. \emph{pairwise error probability}, $\epsilon$ is defined as
\begin{equation} \label{eq:errordef}
    \epsilon = \prob{\hat m(t) \neq m(t)}{} ,
\end{equation}
where $m(t)$ and $\hat m(t)$ are the message sent and estimated at the receiver at time $t$, respectively. This probability has a complex expression in general (not always available in closed-form) that depends on the channel and modulation coding scheme, the transmit power and the channel gain, i.e. $\epsilon \defeq f(\beta,n_c, p_{\text{tx}},h)$. Bounds and approximations in finite block length exist, i.e. when $n_c$ is finite, but remain complex to evaluate \cite{Polyanskiy2010,Anade2020}.

The buffer state evolution, i.e. the number of bits stored in the queue, can be described by a Markov chain with the dynamics
\begin{equation} \label{eq:bufferdyn}
    B(t+1) = \left[B(t) - \beta(t)\cdot \ind{\hat m(t) = m(t)}\right]^+ + N(t) .
\end{equation}
This equation states that the number of bits in the buffer in the next time slot, $B(t+1)$, is the number of bits that is stored at the current time slot, $B(t)$, plus the new bits arriving in the buffer, $N(t)$, minus the number of bits that has been sent through the channel if the transmission is successful, i.e. the indicator function is equal to 1 if so and 0 otherwise. Otherwise the packet remains in the queue and another attempt will occur in the next time slot.

\paragraph{State space.} The state space can be defined as the space containing the channel state, the buffer state and the power consumption state, i.e. $\set{S} = \set{H}\times \set{B}\times \set{P}$.

\paragraph{Action space.} At each time slot, transmitter chooses a power, including the choice not to transmit, i.e. $P_{\text{tx}}(t) = 0$, and a number of bits $\beta(t)$ that is mapped into a codeword\footnote{This transformation is the usual way to consider the encoding phase in information theory. In practice, a transmitter selects a channel coding rate and the resulting bit-train is grouped into symbols in a chosen constellation.} with a fixed block length $n_c$ but with a variable rate $\rho(t) = \beta(t)/n_c$. The action space is then described as $\set{A} = \left\{ 0, p_1, \cdots, p_{\max}\right\}\times \left\{1,\cdots, B(t) \right\}$.

\paragraph{Reward / cost functions.}
In this type of problem, one may be interested to transmit bits with the minimal power while limiting the awaiting time in the buffer. In that case, the global reward can be expressed w.r.t. two cost functions, i.e. the power and the waiting time cost functions \cite{Mastronarde2013,Ngo2010},
$c:\set{A}\times \set{S} \rightarrow \mathbb{R}_{+}$ and $w:\set{A}\times \set{S} \rightarrow \mathbb{R}_{+}$\footnote{Beside the action space, the waiting time cost only depends on the buffer state $\set{B}$ and hence $w$ is incentive to the other states in $\set{S}$.}. The power consumption depends on the transmit power at time $t$ that depends on the target error rate $\epsilon$, the channel state $h$ and the code rate $\rho$, i.e. $\beta$ since $n_c$ is fixed. The power cost can be defined as
\begin{equation} \label{eq:powercost}
    c : \left\{
    \begin{array}{cl}
        \set{A} \times \set{S} & \longrightarrow \mathbb{R}_+ \\
        (a,s) & \longmapsto p_{\text{tot}}(\epsilon,h,\beta)
    \end{array} .
    \right.
\end{equation}
The buffer waiting time cost is defined as
\begin{equation} \label{eq:timecost}
    w : \left\{
    \begin{array}{cl}
        \set{A} \times \set{S} & \longrightarrow \mathbb{R}_+ \\
        (a,s) & \longmapsto \eta \ind{b(t+1) > B_{\max}} + \left(b(t) - \beta (t) \ind{\hat m(t) = m(t)}\right)
    \end{array} ,
    \right.
\end{equation}
The first term represents the cost to be in overflow with $\eta$ a constant, for the sake of simplicity. It means that the cost to pay when the buffer is in overflow is independent of the amount of the overflow\footnote{One can make this cost dependent on the amount of overflow, see \cite{Mastronarde2013}.}. The second term is the holding cost, i.e. the cost for keeping $b - \beta$ bits in the buffer if the transmission is successful.

\paragraph{Policy.} The transmission scheduling policy consists in mapping the system state to an action at each time slot $t$, i.e. according to the buffer state, the channel state observed at the receiver\footnote{The channel state information is assumed to be fed back to the transmitter.} and the target error rate desired, the policy tells us how many information bits stored in the queue we should encode at the next transmission slot and at which power.

Hence a desirable policy should solve an optimization problem. From the cost functions defined previously, the expected discounted power and waiting time costs, given an initial state $s_0\defeq S(0)$, are defined as:
\begin{equation} \label{eq:discpowercost}
    C_{\pi}(s_0) = \Exp{\sum_{t = 0}^{\infty} \gamma^t c\left(A(t),S(t)\right) \left|\right. S(0) = s_0}{\pi} .
\end{equation}
and
\begin{equation} \label{eq:discwaitcost}
    W_{\pi}(s_0) = \Exp{\sum_{t = 0}^{\infty} \gamma^t w\left(A(t),S(t)\right) \left|\right. S(0) = s_0}{\pi} .
\end{equation}
The expectation is taken over the distribution of the policy and the dynamic of the underlying MDP. The problem of finding the minimal power consumption while limiting the waiting time cost can be formally described as \cite{Mastronarde2013,Ngo2010}\footnote{Note that one could have searched for minimizing the waiting time cost under a total power budget as studied in \cite{Goyal2003}, that leads to equivalent strategy.}
\begin{equation}
    \underset{\pi \in \Phi}{\min} \mbox{ }C_{\pi}(s_0) \text{ s.t. } W_{\pi}(s_0) \leq \delta \mbox{ }\forall s_0\in \set{S} .
\end{equation}
The problem relies to a constrained optimisation problem with unknown dynamics. One can combine the power and waiting time cost functions $c$ and $w$ in \eqref{eq:powercost} and \eqref{eq:timecost}, respectively, in a dual Lagrangian expression such that
$\ell(a,s ; \lambda) = c(a,s) + \lambda w(a,s)$, $\lambda \in \mathbb{R}_+$ as proposed in \cite{Mastronarde2013,Ngo2010}. One can hence write an expected discounted Lagrangian cost on the same model than in \eqref{eq:discpowercost}, for instance, but substituting $c$ by $\ell$.

One can apply the Q-learning algorithm detailed in Section \ref{sec:qsarsa} by replacing the reward $R$ in \eqref{eq:qtarget} by the average discounted Lagrangian cost to obtain the optimal policy that minimizes the average power consumed under a buffer delay constraint.

\begin{remark}
The naive implementation of the Q-learning algorithm may be inefficient in terms of the algorithm's convergence time as reported in \cite{Mastronarde2013}. Indeed, Q-learning does not assume any knowledge about the dynamic of the underlying MDP. Hence, the exploration part, which is fundamental in Q-learning, slows down the convergence time due to the large number of combination of states and actions. However, in wireless communication some dynamics may not be completely unknown. The authors in \cite{Mastronarde2013} proposed to use the concept of the post-decision states, presented in wireless communication literature in \cite{Salodkar2008}. The concept consists of reducing the amount of state to explore to take good decisions on the long run by basing the actions to take on states that would be observed considering only the known dynamics.
\end{remark}

\subsection{Example with Deep-RL} \label{sec:drlex}
When the state and action spaces become large, the tabulated methods for SARSA or Q-learning are no longer practical. In that case, methods relying on the approximation of the Q function are meaningful like the linear approximation and those based on deep neural networks.

\subsubsection{Cache enabled communications} \label{sec:linearrl}

Mobile Internet allows anyone to access heterogeneous data in mobility. However, all the contents are not requested the same by the users, i.e. the data do not have the same popularity and some videos, for instance, may be more requested than other files by a user. In order to reduce the data traffic on the backhaul link, and hence the network operating costs, the most requested files can be kept into the storage unit of the base station; this is what is called caching. Hence, the most "popular" files are stored at the base station and can be delivered quickly to the user when requested and hence reducing the cost to download the file from a distant server.

The problem of learning the optimal caching strategy for satisfying users demand or data offloading in various environments has been addressed in lot of works, e.g. \cite[Table V]{Luong2019}. In this section, we will focus on a simple example to illustrate how the caching problem may be addressed with deep Q-learning. The example is an adaptation of \cite{Sadeghi2018} where this problem is studied in a more complex setting. We briefly summarize how to properly choose the action and state spaces and rewards in order to find the optimal caching strategy.

Let us consider a network with a single cell serving many users. Each user may ask for a file $f$ in the set $\set{F} = \left\{1, \cdots, F \right\}$ and we assume that only $M\ll F$ files can be stored at the base station. The files are requested randomly according to a certain distribution characterizing their \emph{popularity} at time $t$. The popularity of the files is modeled as a random vector $\boldsymbol{\mathrm{P}}(t) = \left[\mathrm{P}_1(t), \cdots, \mathrm{P}_F(t)\right]^T$ where the distribution of each popularity can be modeled with Zipf's law\footnote{Zipf's law has been first used to characterize the frequency of occurrence of a word according to its rank.}. This law gives the average number of occurrences of each file and can be estimated online

The goal of the network is to decide the files to cache and those to remove from the cache storing unit at each time slot. Because of the large number of possible requested files the number of possible choices is huge, i.e. $2^M$ with $M\gg 1$. Classical tabulated Q-learning approach, like the example presented in Section \ref{sec:qlearex}, is not suitable.

\paragraph{Action space.} Let $\set{A}$ be the set of \emph{caching action vectors}, that contains the binary action vector $\boldsymbol{A}(t)$ at time $t$ such that $\set{A} = \left\{ \boldsymbol{A} \left|\right. \boldsymbol{A}\in \left\{0,1\right\}^F, \boldsymbol{A}^T \boldsymbol{1} = M\right\}$. $A_f(t)\in \left\{0,1\right\}$, $f\in \left\{1, \cdots, F\right\}$ is a random variable that is equal to 1 if the file $f$ is cached in the BS and 0 otherwise at time $t$.

\paragraph{State space.} The state space is made of the popularity profile and the caching action vector, the latter being also an indicator of the cache status at time $t$. The popularity profile vector $\boldsymbol{\mathrm{P}}$ is assumed to evolve according to a Markov chain with $\left|\set{P}\right|$ states taken in the set $\set{P} = \left\{\boldsymbol{\mathrm{P}^1, \cdots, \boldsymbol{\mathrm{P}^{\left|\set{P}\right|}}}\right\}$. The state space is hence $\set{S} = \set{P}\times \set{A}$.

\paragraph{Reward/cost function.} Similar to the previous example, the reward takes the form of a cost function as
\begin{equation} \label{eq:costcaching}
    c : \left\{
    \begin{array}{cl}
        \set{A} \times \set{S} & \longrightarrow \mathbb{R}_+ \\
        (\vec{a},\vec{p}) & \longmapsto \lambda_1 \vec{a}(t)^T \left( \vec{1} - \vec{a}(t-1)\right) + \lambda_2 \left( \vec{1} - \vec{a}(t)\right)^T \vec{p}(t)
    \end{array} .
    \right.
\end{equation}
The cost function is made of two parts: i) a term related to the cost of not having a file cached in the previous time slot, i.e. term $1-\vec{a}(t-1)$, which needs to be cached in the current time slot and ii) an immediate cost for non caching the file with high popularity profile at time $t$. The constants $\lambda_1$ and $\lambda_2$ allow to balance the importance of these two costs.

\paragraph{Policy.} The goal of the reinforcement learning is to learn the optimal policy $\pi^* : \set{S} \rightarrow \set{A}$ that minimizes the long term weighted average cost function
\begin{equation} \label{eq:policyopt}
    \pi^* = \underset{\pi \in \Phi}{\arg\min} \mbox{ }\Exp{\sum_{t=0}^{\infty} \gamma^t c\left(\left(\vec{A}(t), \vec{P}(t)\right)\right) \left|\right. \vec{S}(0) = \vec{s}(0)}{\pi} ,
\end{equation}
where the expectation is carried out through the distribution of the policy (if stochastic policy is used) and the distribution of the random variables $\vec{A}(t)$ and $\vec{P}(t)$. Given a state $\vec{s}(t)$ at time $t$, the policy looks for the set of files to be stored at time $t+1$, i.e. $\vec{a}(t+1)$ according to the popularity profile observed so far.

By denoting the state transition probabilities as $P_{\vec{S}(t)\left|\right. \vec{S}(t-1) , \vec{A}(t-1)} \defeq p(\vec{s}'\left|\right. \vec{s} , \vec{a})$, the Q-function can be obtained using the Bellman's equation as
\begin{equation} \label{eq:cachebllq}
    q_{\pi}(\vec{s},\vec{a}) = \overline{c(\vec{s},\vec{a})} + \gamma \sum_{\vec{s'}\in \set{S}} p\left(\vec{s'}\left|\right. \vec{s},\vec{a}\right) q_{\pi} (\vec{s'},\vec{a})
\end{equation}
where $\overline{c(\vec{s},\vec{a})} = \Exp{c\left( \vec{A}(t), \vec{P}(t)\right)}{\pi P_{\vec{S'}\left|\right. \vec{S} \vec{A}}}$. Finding the optimal policy in \eqref{eq:policyopt} that is the solution of \eqref{eq:cachebllq}, which is obtained after policy evaluation and improvement steps, requires one to know the dynamics of the underlying Markov chain. The authors in \cite{Sadeghi2018} proposed a Q-learning algorithm with linear function approximation as introduced in Section \ref{sec:deeprl} in order to cope with the high dimensionality of the problem, i.e. $q\left(\vec{s},\vec{a}\right) \approx \boldsymbol{\psi(s)}^T\left( \vec{1} - \vec{a}\right)$ where $\boldsymbol{\psi(s)}$ is a state dependent feature vector that can be expressed as $\boldsymbol{\theta}^{\vec{p}} + \theta^R \vec{a}$ in which $\boldsymbol{\theta}^{\vec{p}}$ represents the average cost of non caching files when the popularity is in state $\vec{p}$, and $\theta^R$ is the average cache refreshing cost per file. By adapting the recursion in \eqref{eq:sgdq} to the problem above, i.e. the reward is replaced by the cost function in \eqref{eq:costcaching}, the maximization is replaced by a minimization operation. One can show that this technique is able to converge to the optimal caching strategy for large number of files and popularity profiles.

\subsubsection{Multi-user scheduling and power allocation} \label{sec:dqnrl}

Let us extend the problem presented in Section \ref{sec:qlearex} by adding multiple users in the system. An orthogonal frequency division multiple access downlink network, with a single cell and $K$ users is considered. The whole time-frequency resource is divided in $N_{\text{rb}}$ resource blocks (RBs) and one RB is made of $N_s$ subcarriers. The base station handles $K$ queues, one for each user, and has to serve the users by allocating the suitable transmission powers and the number of RBs in order to transmit the maximum number of bits with the minimal total power consumption and under buffer waiting time constraints.

The channel gain is considered to be constant on one RB, i.e. over $N_s$ subcarriers and during $\Delta T_{\text{RB}}$, and varies from one RB to another according to the distribution
$$P_{\vec{H}(t+1) \left|\right. \vec{H}(t)} (\vec{h'} \left|\right. \vec{h}) = \prod_{k=1}^K \prod_{r=1}^{N_{\text{rb}}} P_{H_{kr}(t+1)\left|\right. H_{kr}(t)} (h'_{kr}\left|\right. h_{kr})$$
with $H_{kr}(t) \in \mathcal{H}$ the random variable representing the channel gain of user $k\in \mathcal{K} = \left\{1, \cdots, K\right\}$ on RB number $r \in \mathcal{N}_{\text{RB}} = \left\{1,\cdots, N_{\text{RB}}\right\}$ at time $t$. This relation means that the channel gains are independent from one RB to another in frequency and from a user to another. Similarly than in Section \ref{sec:qlearex}, the application layer of each user generates $N_k(t)$ bits/packets at time slot $t$ according to the distribution $P_N(n)$. The generated bits are stored in a buffer for each user characterized by its size $B_k(t) \in \mathcal{B}_k$, where $\mathcal{B}_k$ is defined similarly than in Section \ref{sec:qlearex} for all users $k \in \mathcal{K}$. We assume that only a packet of $L$ information bits can be sent per user and per time slot. BS can choose the modulation and coding scheme (MCS) for each user, $\text{mcs}_k \in \mathcal{MC} = \left\{\text{mcs}_1, \cdots , \text{mcs}_C\right\}$, i.e. a couple $\left(\chi_k , \rho_k\right)$ where $\chi_k$ and $\rho_k$ are the modulation order in a QAM constellation and the rate of channel encoder for user $k$, respectively. The MCSs are ordered from the lowest to the highest spectral efficiency. A set of MCS used for the LTE system can be found in \cite[Table I]{Maaz2017}.

The power and RBs allocation can be done at once by the BS by choosing the transmission power vector to allocate to user $k$ over all the RBs at time $t$, $\vec{P}_k(t)$, in the power state space $\mathcal{P}$ such that $\mathcal{P} = \left\{ \vec{P} \left|\right. \vec{P}\in \left\{0, p_1, \cdots, p_{\max}\right\}^{N_{\text{RB}}}, \frac{1}{N_{\text{RB}}} \vec{P}^T \vec{1}_{N_{\text{RB}}}  \leq \overline{p_{\text{tot}}} \right\}$, where $\overline{p_{\text{tot}}}$ is the maximum average power budget that can be allocated to a user over all the subcarriers. The power of user $k$ on RB $r$ at time $t$, i.e. $p_{kr}(t)$, is null if the RB is not used by the user. In papers where this kind of problem is handled with classical convex optimization tools, RB allocation is dealt with an auxiliary variable that is equal to one when user $k$ uses RB $r$ and 0 otherwise \cite{Hui2007,Maaz2017}.

The error probability of user $k$ is defined as in \eqref{eq:errordef}, i.e. $\epsilon_k \defeq \prob{\hat \omega_k(t) \neq \omega_{k}(t)}{}$, where $\omega_k(t)$ is the message sent by user $k$ at time $t$. It depends on the chosen MCS, the transmission power and the channel state experienced over each RB allocated to the user $k$. The queue dynamic of user $k$ is then
\begin{equation}
    B_k(t+1) = \left[B_k(t) - L \cdot \ind{\hat \omega_k(t)} = \omega_k(t)\right]^+ + N_k(t).
\end{equation}

\paragraph{State space.} The state space is made of the buffer state of each user, the channel gain state of each user on each RB allocated to it and the power consumed by each user, i.e. $\mathcal{S} = \mathcal{H}^{ K N_{\text{RB}}} \times \mathcal{B}^{ K} \times \mathcal{P}^{ K }$.

\paragraph{Action space.} BS chooses the power and the MCS couple $\text{mcs}_k = (\chi_k,\rho_k)$ to allocate to all users. The action space is hence $\mathcal{A} = \mathcal{P}^K \times \mathcal{MC}^K$.

\paragraph{Reward/cost functions.} The objective of the network operator may be to minimize the power consumed to serve all the $K$ users in the cell while guaranteeing a limited buffer waiting time. We assume that the power consumption is only made of the transmit power, i.e. static power consumption is neglected. The power consumed by user $k$ depends on the target error rate required, the observed channel state and the used MCS. The total power consumption of the cell can be written as
\begin{equation} \label{eq:powercosttot}
    c : \left\{
    \begin{array}{cl}
        \set{A} \times \set{S} & \longrightarrow \mathbb{R}_+ \\
        (\vec{a},\vec{s}) & \longmapsto \sum_{k=1}^K \vec{p}_k(\epsilon_k,\vec{h}_k,\text{mcs}_k)^T \vec{1}_{N_{\text{RB}}}
    \end{array} .
    \right.
\end{equation}
The buffer waiting time cost is defined similarly than in \eqref{eq:timecost}, hence the total cost of the waiting time over the cell is
\begin{equation} \label{eq:timecosttot}
    w : \left\{
    \begin{array}{cl}
        \set{A} \times \set{S} & \longrightarrow \mathbb{R}_+ \\
        (\vec{a},\vec{s}) & \longmapsto \sum_{k=1}^K \eta \ind{b_k(t+1) > B_{\max}} + \left(b_k(t) - L \ind{\hat \omega_k(t) = \omega_k(t)}\right)
    \end{array} ,
    \right.
\end{equation}
The average Lagrangian discounted cost, which is identified to our Q-function, can be obtained similarly than in Section \ref{sec:qlearex}.
\begin{equation} \label{eq:qcostdrl}
    q_{\pi}\left(\vec{s}_0,\vec{a}_0 ; \lambda\right) = \Exp{\sum_{t=0}^{\infty} \gamma^t \ell\left( \vec{A}(t) , \vec{S}(t) ; \lambda \right) \left|\right. \vec{S}(0) = \vec{s}_0 , \vec{A}(t) = \vec{a}_0}{\pi} ,
\end{equation}
where $\vec{s}_0, \vec{a}_0$ are the initial state and action vectors and $\ell\left(\vec{a}, \vec{s} ; \lambda\right)$ is defined similarly than in Section \ref{sec:qlearex} but with the functions in \eqref{eq:powercosttot} and \eqref{eq:timecosttot} and $\lambda$ is the Lagrange multiplier.

\paragraph{Policy.} The problem is to find the policy $\pi^*$ that minimizes \eqref{eq:qcostdrl} for all $\left( \vec{s}_0 , \vec{a}_0\right)$ for a given $\lambda$. This optimization problem deals with a huge number of variables and the classical tabulated Q-learning requires to much time to converge and too much capacity storage. In this situation, DRL can be a suitable solution to approach the function in \eqref{eq:qcostdrl} by implementing, for instance, an FNN as illustrated in Fig. \ref{fig:fnn}. For a given set of parameters of the deep network at time $t$, i.e. $\vec{\Theta}_t$, the loss function is defined as in \eqref{eq:drlloss} where $R(t) = \ell\left(\vec{A}(t) , \vec{S}(t) ; \lambda\right)$ and the optimal Q-function is approached at time $t$ by $q_{\vec{\Theta}_t}\left(\vec{S}(t) , \vec{A}(t) ; \lambda \right)$. The optimal set of weights of the neural network can be updated using \eqref{eq:sgddeep} with proper variables.

\subsection{Examples with MAB} \label{sec:mabex}

\subsubsection{Multi-channel access problem} \label{sec:mabosa}

Let us consider a set of $K$ independent channels, i.e. $\set{K}=\left\{1,\cdots,K\right\}$ that can be used opportunistically by $U$ users, with $K\geq U$ to communicate with a BS. The channels may be also used by other users that belong to a primary network. The users can sense one or more channels to estimate if they are used or not. To perform this task, the users can rely on multiple signal processing techniques ranging from the simple energy detector to sophisticated signal classifiers \cite{Axell2012}. If the channel is detected to be free, the user transmits in that band, otherwise the transmitter remains silent. The action space is hence $\set{A} = \{\text{transmit},\text{silent}\}$. When the channel is free however, it may be rated with a low or a high quality depending on the level of the received SINR for instance or the actual data rate a user experienced on it. The state space may be limited to $\set{S} = \left\{ \text{busy}, \text{free}\right\}$ but the quality of the band can be included in the reward function as it has been proposed in \cite{Oksanen2012, Oksanen2015}. In the case where a single user is considered i.e. $U=1$:
\begin{equation}
    R_k(t) = (1 - S_k(t)) f(S_k(t)),
\end{equation}
where $S_k(t)\in \left\{0,1\right\}$ is the state of band $k$ at time $t$ where $1$ means that the band is detected as occupied and $0$ that is free. Moreover, $f(S_k(t))$ is the observed data rate on band $k$ when it is in the state $S_k(t)$, and it can be considered that $f(S_k(t)) \in \left[0,1\right], \forall S_k(t)$\footnote{By convention, $f(1) = 0$. Moreover, the experienced data rate can be normalized w.r.t. the channel capacity achievable in that channel.}.

The goal for an agent is to select the band $k$ that maximizes the data rate on the long run. If the expected rewards of each band were known, the optimal strategy would be to sense and to transmit (when possible) always on the band presenting the maximal expected reward. In the absence of this knowledge, the agent has to \emph{learn} what are the best bands and concentrate on them. An index-based policy can be proposed to solve this problem, like UCB algorithm which is order-optimal as explained in Section \ref{sec:mab}. For instance, in \cite{Oksanen2015}, authors proposed to compute the index $I_k(t)$ for each band $k$ and choose the one with the highest value at the next round. The index is computed as
\begin{equation}
    I_k(t) = \overline{r_k(t)} + g\left(\frac{t}{n_k(t)}\right) ,
\end{equation}
where $n_k(t)$ is the number of times band $k$ has been sensed up to time $t$, $\overline{r_k(t)} = \frac{1}{n_k(t)} \sum_{t' = 1}^{n_k(t)} r_k\left(p_k(t')\right)$ where $p_k(t')$ is a sensing time instant corresponding to the $t'$-th visit of the band $k$. Finally, $g$ is a concave function with $g(1) = 0$, e.g. $g(x) = \sqrt{\log(x)}$. One may notice that this function is different from the classical bias of UCB introduces in Section \ref{sec:mab}, which is in the form $\sqrt{\frac{\log(t)}{n_k(t)}}$. The authors proved that the proposed index policy has an aggressive exploration characteristic compared to the UCB where the bias increases slowly. An aggressive exploration statistic means that the optimal data rate will be reached faster than in a non aggressive exploration characteristic (classical UCB) but it will also be lower than classic UCB.

According to the statistical model of the rewards, the problem described may fall into the classes of rested MAB or restless MAB. If the rewards, i.e. the experienced data rate of the users, are i.i.d. on each band, the problem is a rested-MAB. If however, the state $S_k(t)$ is described by a 2-state Markov chain, i.e. Gilbert-Elliot model, so the data rate, and hence the problem can be classified as a restless MAB. In \cite{Oksanen2015} authors showed that the policy is order-optimal for i.i.d. and Markovian rewards as well. 

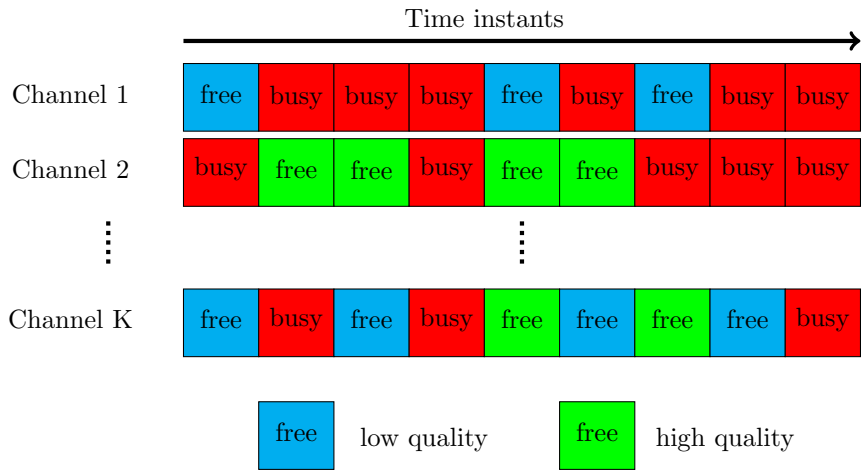
\begin{figure}[htbp]
\begin{center}
    \begin{tikzpicture}
        \draw[ultra thick , ->] (-4,1.2) -- (5,1.2);
        \node at (0,1.5){Time instants};
        \node at(-5.5,0.5){Channel 1}; 
        \draw[fill=cyan] (-4,0) rectangle (-3,0.9) ; 
        \node at(-3.5,0.5){free}; 
        \draw[fill=red] (-3,0) rectangle (-2,0.9); 
        \node at(-2.5,0.45){busy}; 
        \draw[fill=red] (-2,0) rectangle (-1,0.9); 
        \node at(-1.5,0.45){busy}; 
        \draw[fill=red] (-1,0) rectangle (0,0.9); 
        \node at(-0.5,0.45){busy}; 
        \draw[fill=cyan] (0,0) rectangle (1,0.9); 
        \node at(0.5,0.5){free}; 
        \draw[fill=red] (1,0) rectangle (2,0.9); 
        \node at(1.5,0.45){busy}; 
        \draw[fill=cyan] (2,0) rectangle (3,0.9); 
        \node at(2.5,0.5){free}; 
        \draw[fill=red] (3,0) rectangle (4,0.9); 
        \node at(3.5,0.45){busy}; 
        \draw[fill=red] (4,0) rectangle (5,0.9); 
        \node at(4.5,0.45){busy}; 
        \node at(-5.5,-0.5){Channel 2}; time frame of channel 2
        \draw[fill=red] (-4,-1) rectangle (-3,-0.1) ; 
        \node at(-3.5,-0.5){busy}; 
        \draw[fill=green] (-3,-1) rectangle (-2,-0.1); 
        \node at(-2.5,-0.5){free}; 
        \draw[fill=green] (-2,-1) rectangle (-1,-0.1); 
        \node at(-1.5,-0.51){free}; 
        \draw[fill=red] (-1,-1) rectangle (0,-0.1); 
        \node at(-0.5,-0.5){busy}; 
        \draw[fill=green] (0,-1) rectangle (1,-0.1); 
        \node at(0.5,-0.51){free}; 
        \draw[fill=green] (1,-1) rectangle (2,-0.1); 
        \node at(1.5,-0.51){free}; 
        \draw[fill=red] (2,-1) rectangle (3,-0.1); 
        \node at(2.5,-0.5){busy}; 
        \draw[fill=red] (3,-1) rectangle (4,-0.1); 
        \node at(3.5,-0.5){busy}; 
        \draw[fill=red] (4,-1) rectangle (5,-0.1); 
        \node at(4.5,-0.5){busy}; 
        \draw[dotted, ultra thick] (0.5 , -1.2) -- (0.5 , -1.8); 
        \draw[dotted, ultra thick] (-5 , -1.2) -- (-5 , -1.8); 
        \node at(-5.5,-2.5){Channel K}; 
        \draw[fill=cyan] (-4,-3) rectangle (-3,-2.1) ; 
        \node at(-3.5,-2.5){free}; 
        \draw[fill=red] (-3,-3) rectangle (-2,-2.1); 
        \node at(-2.5,-2.5){busy}; 
        \draw[fill=cyan] (-2,-3) rectangle (-1,-2.1); 
        \node at(-1.5,-2.5){free}; 
        \draw[fill=red] (-1,-3) rectangle (0,-2.1); 
        \node at(-0.5,-2.5){busy}; 
        \draw[fill=green] (0,-3) rectangle (1,-2.1); 
        \node at(0.5,-2.5){free}; 
        \draw[fill=cyan] (1,-3) rectangle (2,-2.1); 
        \node at(1.5,-2.5){free}; 
        \draw[fill=green] (2,-3) rectangle (3,-2.1); 
        \node at(2.5,-2.5){free}; 
        \draw[fill=cyan] (3,-3) rectangle (4,-2.1); 
        \node at(3.5,-2.5){free}; 
        \draw[fill=red] (4,-3) rectangle (5,-2.1); 
        \node at(4.5,-2.5){busy}; 
        
        \draw[fill=cyan] (-3 , -4.5) rectangle (-2 , -3.6);
        \node at(-2.5 , -4){free};
        \node at( -0.8, -4.1){low quality};
        
        \draw[fill=green] (1 , -4.5) rectangle (2 , -3.6);
        \node at(1.5 , -4){free};
        \node at( 3.2, -4.1){high quality};
   \end{tikzpicture}
\end{center}
\caption{Dynamic spectrum access with different channel qualities} \label{fig:osaqual}
\end{figure}

In the previous example, the reward signal was a function of the state of the channel sensed and the data rate experienced in this channel by the user. In some applications, one may be interested in acquiring knowledge both on the channel availability and on channel quality. The channel quality has to be taken in a broad sense. It may be the noise level, including the average power density of the interference, the average data rate experienced in this band, etc. The dynamic spectrum access with different channel quality can be represented as in Fig. \ref{fig:osaqual}. In this example, the state space is extended to $\set{S} = \left\{\text{busy},\text{low quality}, \text{high quality}\right\}$. The action space is still the same, i.e. $\set{A} = \{\text{transmit},\text{silent}\}$, but the transmission now occurs on a state, i.e. channel is free, that can be explicitly rated with a high or low quality. On may solve this problem by considering a full RL approach, using value function as described in Section \ref{sec:rltheory}. However, a simple index-based policy can perform very well in this context. Authors in \cite{Navik_2017} proposed that each user $u\in \left\{1,\cdots,U\right\}$ can compute the following index for each band and choose the maximum one as the next band to sense
\begin{equation} \label{eq:inducbnavik}
    I^u_k(t) = \frac{1}{n^u_k(t)} \sum_{t' = 1}^{n_k^u(t)} s_k^u(p_k^u(t')) - q^u_k(t) + \sqrt{\frac{\alpha \log(t)}{n_k^u(t)}}
\end{equation}
where the first term is the empirical average of the reward obtained by user $u$ when sensing band $k$, i.e. 0 if the channel is occupied and 1 if the channel is free. The second term is a function of the empirical average of the quality of band $k$ for user $u$ and is expressed as
\begin{equation} \label{eq:Q}
    q_k^u(t) = \frac{\beta m_k^u(t) \log(t)}{n_k^u(t)} ,
\end{equation}
where $m_k^u(t) = g_{*}^u(t) - g_1^{u,k}(t)$ and $g_1^{u,k}(t)$ denotes the empirical average of the quality of band $k$, when available, sensed by user $u$ at time $t$ and is expressed as
\begin{equation} \label{eq:quality}
    g_1^{u,k}(t) = \frac{1}{n_k^u(t)} \sum_{t'=1}^{n_k^u(t)} r_1^{u,k}(p_k^u(t')) ,
\end{equation}
where $r_1^{u,k}(p_u^k(t'))$ is the reward obtained by user $u$ rating the quality of band $k$, between 0 and 1, at time $p_k^u(t')$. Finally, $g_*^u(t) = \max_{k\in \set{K}} g_1^{u,k}(t)$. The last term in \eqref{eq:inducbnavik} is the classical exploration term of UCB algorithm where the parameter $\alpha$ forces the exploration of other bands to find channels that are the most often available.

The parameter $\beta$ in \eqref{eq:Q} gives weight to the channel quality. At each of iteration, the agent computes the empirical mean, up to the current time instant, on the quality observed if the band is free. In the same time, the best channel among those already tried is updated and the score is computed by weighting the difference between the estimated best channel and the current channel. If $\alpha$ and $\beta$ increase, the agent explores more than it exploits. When $\alpha$ and $\beta$ decrease, the empirical mean of the states dominates the index calculation and the exploitation of the best band computed in the previous iteration is favored.

Due to the restless nature of the problem, the index computation cannot be done when one just starts observing a Markov chain after being selected. Indeed, arms evolve independently irrespective of which band is selected or not (i.e. \emph{restless} MAB) and the distribution of rewards that user $u$ gets from a band $k$ is a function of the time elapsed since the last time user $u$ sensed this band. The sequence of observations of a band that is not continuously sensed does not correspond to a Markov chain. To overcome this issue, when user $u$ observes a given band, algorithm waits for encountering a predefined state, named \emph{regenerative state} e.g. $\xi_k$, \cite{tekin_2011}. Once $\xi_k$ is encountered, rewards are started to be recorded until $\xi_k$ be observed a second time and the policy index $I_k^u$ is computed and another band selected according to the result. This structure is necessary to deal with the restless nature of the problem in order to re-create the condition of continuous observation of Markov chain. It is worth mentioning, however, that exploitation in this context occurs whenever a free band is detected. 

Moreover, the multi-player setting makes the problem rather involved since collisions between agents need to be handled. The random rank idea from \cite{Anandkumar_2011} has been adapted to this problem. Each user maintains an ordered set of channel indexes (arms indexes), i.e. $\set{K}_u = \sigma_u \left( \set{K} \right)$, where $\sigma_u$ is a permutation of $\left\{1, \cdots, K\right\}$ for user $u$, with $\sigma_u(1) > \cdots > \sigma_u(K)$ from the best to the worst rated. The rank $r$ for user $u$ corresponds to the $r-$th entry in the set $\set{K}_u$. If users $u$ and $u'$ choose the same channel to sense the next time slot, they collide. In that case, they draw a random number from their respective sets $\set{K}_u$ and $\set{K}_{u'}$ as their new rank and go for these new channels in the next time slot.

\subsubsection{Green networking} \label{sec:mabgreennet}

The radio access networks are not used at their full capacity all the time. Experimental results in voice call information recorded by operators over one week exhibit periods with high traffic load and others with moderate to low traffic \cite{Oh_2013}. Hence, it may be beneficial for a network operator to dynamically switch off BS that does not handle high traffic in its cell at a given time in order to maximize the energy efficiency of the network. However, the set of BS to be switched OFF should be chosen with care while maintaining a sufficient quality of service for users.

Let us consider an heterogeneous wireless cellular network made of macro and small cells where the set of BS $\set{Y} = \{1,2,\cdots,Y\}$ lies in a two dimensional area in $\mathbb{R}^2$, each serving a cell $k$.  The decision to switch ON or OFF a BS is taken by a central controller and depends on the traffic load of each cell and its power consumption. The traffic load $\rho_k(t)$ of a cell $k$ at time $t$ depends on the statistic of the arrival and departure processes and on the data rate $\Theta_k(x,t) $ that can be provided by cell $k$ to the user positioned at
$x$ at time $t$.

The maximization of the energy efficiency of the network by selecting the set of transmitting BS is an NP-hard problem and can be expressed as following \cite{Navik2019}:
\begin{equation}\label{eq:max_ee_mab}
\begin{array}{lll}
\set{Y}_{\text{on}^*}\left(t\right) = & \arg \underset{\set{Y}_{\text{on}}\left(t\right)}{\max} \left[\sum\limits_{k \in \set{Y}_{\text{on}}\left(t\right)}\frac{\sum\limits_{x \in \set{C}_k\left(t\right)}\Theta_k(x,t)}{P_k(t)}\right] \mbox{ s.t. } & \\
(c_1) & 0 \leq \rho_k(t) \leq \rho_{\text{th}}, \forall k \in \set{Y}_{\text{on}}\left(t\right) & \\
(c_2) & \Theta_k(x,t) \geq \Theta_{\min}, \forall x \in \set{C}_k\left(t\right), \forall k \in \set{Y}_{\text{on}}\left(t\right) & \\
(c_3) & \set{Y}_{\text{on}}\left(t\right) \neq \emptyset &
\end{array}
\end{equation}
where $\set{Y}_\text{on}(t)$ is the set of active BS at time $t$, $P_k(t)$, $\set{C}_k(t)$ are the power consumed and the coverage of cell $k$ at time $t$, respectively. Moreover, $\rho_{\text{th}}$ and $\Theta_{\min}$ are the traffic load upper limit and the minimum required data rate per user, respectively. Constraint $(c_1)$ is stated for stability reason\footnote{A traffic load that is too high results in diverging queue size in the network.}, $(c_2)$ states that each user has to be served with a minimum data rate and $(c_3)$ ensures that at least one BS is active at each time slot. Finding the optimal configuration by an exhaustive search would be prohibitive in large networks since the optimal BS active set belongs to a set of $2^Y-1$ combinations.

\begin{figure}[htbp]
\begin{center}
    \begin{tikzpicture}
        \draw (-2,0) rectangle (0,1) ; 
        \draw (-2,-6) rectangle (2,-3); 
        \draw (1 , -3.8) circle (0.4) ;
        \draw (0.7 , -5.2) circle (0.4) ;
        \draw (-1.1,-3) -- (-0.2,-3.9) ;
        \draw (-0.2,-3.9) -- (-0.5,-4.5) ;
        \draw (-0.5,-4.5) -- (-2,-4.2) ;
        \draw (-0.5,-4.5) -- (-0,-6) ;
        \draw (-0.2,-3.9) -- (2,-5) ;
        \draw[red] (-0.1,-3.4) -- (0.1,-3.6) ; \draw[red] (-0.1,-3.6) -- (0.1,-3.4) ; 
        \draw[red] (-1,-4) -- (-0.8,-4.2) ; \draw[red] (-1,-4.2) -- (-0.8,-4) ; 
        \draw[red] (-1.3,-5) -- (-1.1,-5.2) ; \draw[red] (-1.3,-5.2) -- (-1.1,-5) ; 
        \draw[red] (0.2,-4.7) -- (0.4,-4.9) ; \draw[red] (0.2,-4.9) -- (0.4,-4.7) ; 
        \draw[blue] (1,-3.7) -- (1,-3.9) ; \draw[blue] (0.9,-3.8) -- (1.1,-3.8) ; 
        \draw[blue] (0.7,-5.1) -- (0.7,-5.3) ; \draw[blue] (0.6,-5.2) -- (0.8,-5.2) ;
        \node at(-1,0.5){Controller}; 
        \node at(-0.5,1.3){RQoS-UCB policy}; 
        \node at(4,0.8){Action $\vec{a}(t) = \left[a_1(t), \cdots, a_Y(t)\right]$} ;
        \node at(4,-0.1){$a_i(t) = \left\{ \begin{array}{ll} 1 & \text{BS } i \text{ active} \\ 0 & \text{BS } i \text{ inactive}\end{array}\right. $} ;
        \node at(-2,-2){$\begin{array}{c} \text{Reward} \\ R(t+1) \end{array}$} ;
        \node at(-5,-2){$\begin{array}{c} \text{State} \\ S(t+1) \end{array}$} ;
        \node at(0,-6.2){Network environment} ;
        \draw[red] (2.4,-5) -- (2.6,-5.2) ; \draw[red] (2.4,-5.2) -- (2.6,-5) ;
        \node at(3.7,-5){\small Macro BS} ;
        \draw[blue] (2.5,-5.5) -- (2.5,-5.7) ; \draw[blue] (2.4,-5.6) -- (2.6,-5.6) ;
        \node at(3.7,-5.6){\small Micro BS} ;
        \draw (0,0.5) -- (6.5,0.5) ; 
        \draw (6.5,0.5) -- (6.5,-4.5) ; 
        \draw[->] (6.5,-4.5) -- (2,-4.5) ; 
        \draw (-2,-4) -- (-3,-4) ; 
        \draw (-2,-5) -- (-4,-5) ; 
        \draw (-3,-4) -- (-3,0.25) ; 
        \draw (-4,-5) -- (-4,0.75) ; 
        \draw[->] (-3,0.25) -- (-2,0.25) ; 
        \draw[->] (-4,0.75) -- (-2,0.75) ;  
    \end{tikzpicture}
\end{center}
\caption{RL framework for BS switching operation} \label{fig:RL_BS}
\end{figure}
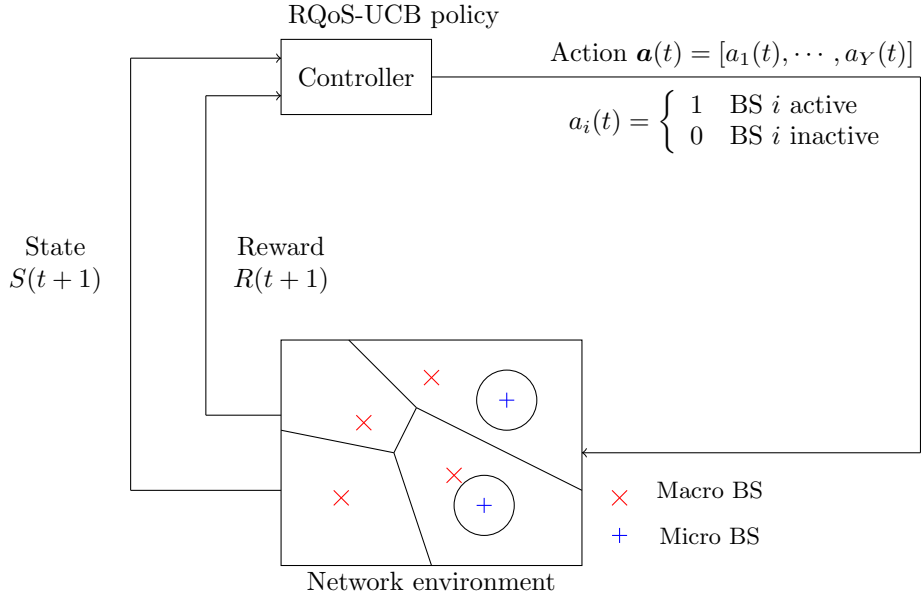

Authors in \cite{Navik2019} have shown that this problem can be solved with MAB formulation. The problem can be illustrated with Fig. \ref{fig:RL_BS} where at each iteration, the central controller chooses an action $\vec{a}$ among $\left|\set{A}\right| = 2^Y - 1$ possible actions, i.e. $\vec{a}(t) = \left[a_1(t), \cdots, a_Y(t) \right]^T$ with $a_k(t) = 1$ if BS $k$ is switched ON at $t$ and 0 otherwise, and where $\set{A}$ is the action space. The state is represented by a random variable $S(t) \in \left\{0,1\right\}$ where $s(t) = 1$ if all constraints in \eqref{eq:max_ee_mab} are satisfied and 0 otherwise. In other words, the value of the state of the Markov chain relies on the fact that the selected action leads to a feasible solution of \eqref{eq:max_ee_mab}. Observing the state $s(t)$ and taking the action $\textbf{a}(t)$ at time $t$ lead to the network in the state $S(t+1)$ and give a reward $R(t+1)$ in the next time slot according to the conditional transition probability distribution introduced in \eqref{eq:srdist}. The reward is the energy efficiency computed as in the cost function in \eqref{eq:max_ee_mab}. 

In \cite{Navik2019} the authors proposed to apply the same policy as in \eqref{eq:inducbnavik}, where the index of the user can be dropped out, $k$ represents the number of the actions and where the middle term $q_k(t)$ is expressed as in \eqref{eq:Q} and where $r_1(t)$ in \eqref{eq:quality} is the energy efficiency of the network at time $t$ when the set of active BS is such that constraints in \eqref{eq:max_ee_mab} are satisfied, i.e. $s(t)=1$.

\begin{remark}
    Of course, modeling \eqref{eq:max_ee_mab} with MAB framework does not change that the problem dimensionality is exponential in the number of BS. However, MAB explores only once all network configurations to assign an index to each of them and then chooses the next configuration according to the highest index computed at the previous iteration with \eqref{eq:inducbnavik}, instead of doing an exhaustive search at each time slot.
\end{remark}

\begin{remark} This problem could also be addressed with DRL technique. Indeed, if the number of BS is too large, the convergence time of MAB algorithms is too large which makes the problem unsolvable for large scale networks. DRL can be used instead and the set of BS to switch on may be obtained by a DNN for each state of the environment.
\end{remark}

\subsection{Real world examples} \label{sec:mabpoc}

We propose to give details on a few concrete implementations of RL for communication, which rely on theoretical applications of RL using the UCB algorithms for single user cognitive radio MAB-modeled problems in \cite{Wassim_2009} and for the opportunistic spectrum access (OSA) scenario in \cite{Wassim_2010}. In these examples, UCB1 \cite{Auer_2002} algorithm is mostly used as a proof of the pertinence of bandit algorithms for free spectrum-access problems, but other bandit algorithms have also been considered and could be implemented as well. 

\begin{figure}[htpb]
	\begin{center}
	\includegraphics[viewport = 30 80 700 450, clip, width=\textwidth]{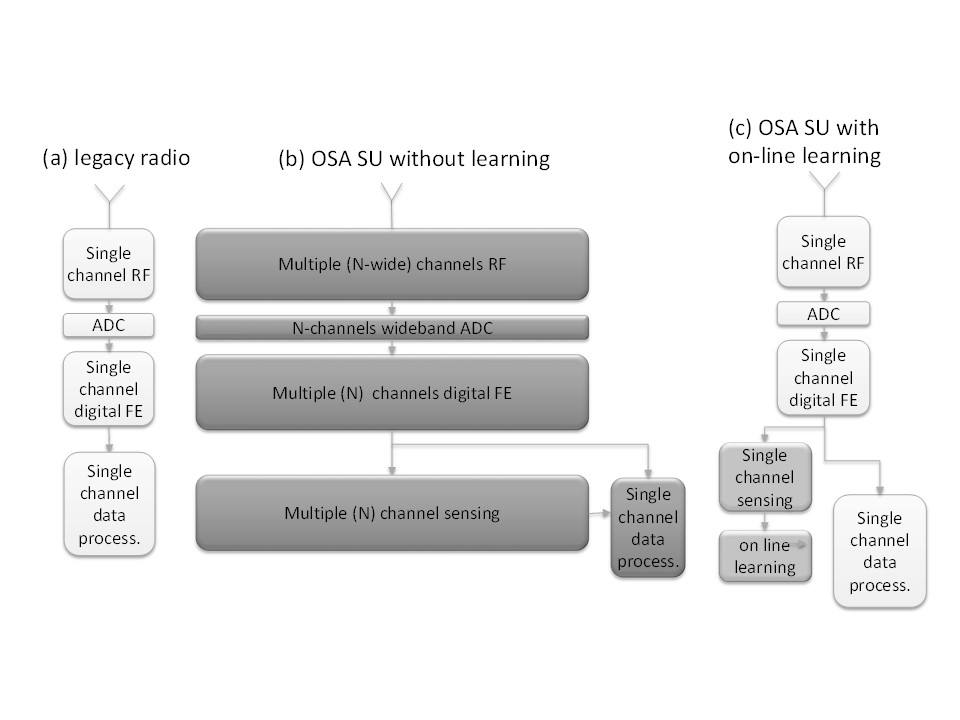}
	\caption{Comparison of the receiver architectures  for OSA with or without RL.} 
	\label{fig:RLarchi_1}
	\end{center}
\end{figure}

In the spectrum access context, the goal is to be able to manage a large spectrum without adding complexity to the radio system by enlarging its bandwidth (and consequently without adding complexity to the transceivers). Figure \ref{fig:RLarchi_1} shows that learning is a mean to decrease the receivers architecture complexity while maintaining a legacy bandwidth to the OSA radio system with extended capabilities. RL enables to reconstruct the global bandwidth knowledge, thanks to successive small scale investigations. RL offers a light solution, in terms of implementation complexity cost compared to wide (full)band OSA systems.  

\subsubsection{Implementation at the post-processing level}

\begin{figure}[htpb]
	\begin{center}
	\includegraphics[viewport = 35 90 670 450, clip, width=\textwidth]{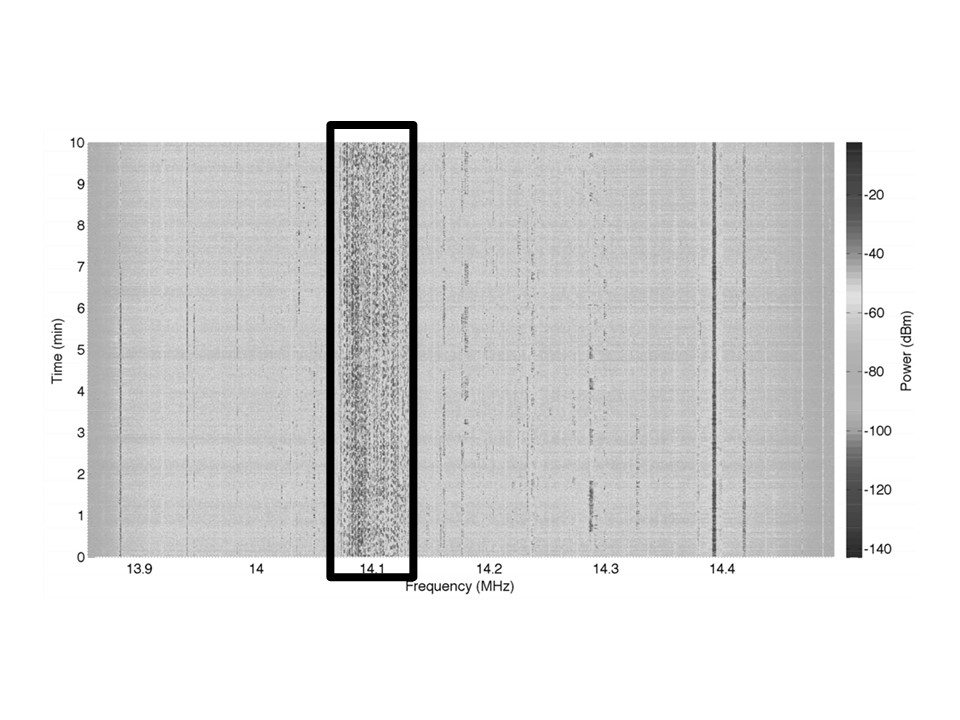}
	\caption{HFSA\_IDeTIC\_F1\_V01 database extraction for HF channel traffic during a radio-ham contest at 14.1 MHz (area highlighted in the rectangle) \cite{Melian_2016}.} 
	\label{fig:HF_1}
	\end{center}
\end{figure}

Due to the unique world wide covering characteristic of high frequency (HF) transmissions, there is a high necessity to reduce collisions between users which act in a decentralized manner. RL algorithms have been applied on real measurements of the HF spectrum that has been recorded during a radio-ham contest \cite{Melian_2016}, i.e. when HF radio traffic is at its highest. Figure \ref{fig:HF_1} shows the subset of the spectrum data the MAB algorithms considered in \cite{Melian_2016}. The goal of learning is to enable users to find yellow unoccupied slots in time and frequency before transmitting and to maximize the probability that no collision occurs during the transmission duration. For instance, the cognitive device should avoid the frequencies delimited by the rectangle in Figure \ref{fig:HF_1}, as it is a highly used bandwidth by primary users.

The solution in \cite{Melian_2016} proposes a new hybrid system which combines two types of machine learning techniques based on MAB and Hidden Markov Models (HMM). This system can be seen as a meta-cognitive engine that automatically adapts its data transmission strategy according to the HF environment's behaviour to efficiently use the spectrum holes. The proposed hybrid algorithm, i.e. combining a UCB algorithm and HMM, allows to increase the time the opportunistic user transmits when conditions are favourable, and is also able to reduce the required signalling transmissions between the transmitter and the receiver to inform which channels have been selected for data transmission. 

Table \ref{tab:SummaryHF_1} sums-up the characteristics of the RL algorithm implementation for this use case. The reward is the channel availability detected by the cognitive users. More details about these measurements can be found in \cite{Melian_2016}.

\begin{table}[hbtp]
\centering 
\begin{tabular}{|c|c|}
\hline
Characteristics & Comment \\
\hline
\hline
RL algorithm &  HMM combined with UCB1   \\
\hline
Reward & channel availability     \\
\hline
Implementation side & HF transceiver   \\
\hline
method for RL feedback loop & Sensing   \\
\hline
\end{tabular}
\caption{Summary for HF signals post-processing.} \label{tab:SummaryHF_1}
\hspace{0in}
\end{table}

\subsubsection{Implementation in a Proof-of-Concept}

First real-time RL implementation on real radio signal took place in 2013 for OSA scenario and was first published in 2014 \cite{Moy_2014} at the Karlshruhe Workshop of Prof. Friedrich Jondral and extended hereafter in \cite{Darak_J_2016}. This consisted of a proof-of-concept (PoC) in the laboratory conditions with one USRP\footnote{\url{https://www.ettusresearch.com/}} platform emulating the traffic generated by a set of primary users, i.e. users that own the frequency bands, and another USRP platform running the sensing and learning algorithm of one secondary user, i.e. user that opportunistically exploits the licensed band. Both i.i.d and Markovian MAB traffic models have been tested. UCB1 algorithm was used first in order to validate the RL approach, but then other bandit algorithms have been implemented later, e.g. Thompson Sampling, KL-UCB. The multi-user version has been implemented in \cite{Darak_2016}, moreover, several videos implementing the main UCB algorithms in a USRP-based platform demonstrating the real-time learning evolution under various traffic models (i.i.d. and Markovian) can be found on Internet\footnote{\url{https://www.youtube.com/channel/UC5UFCuH4jQ\_s\_4UQb4spt7Q/videos}}. In order to help the experimental community to verify and develop new learning algorithms, an exhaustive Python code library and framework for simulations have been provided on GithHub\footnote{“SMPyBandits: an Open-Source Research Framework for Single and Multi-Players Multi-Arms Bandits (MAB) Algorithms in Python”} \footnote{code on \url{https://GitHub.com/SMPyBandits/SMPyBandits}} \footnote{documentation on \url{https://SMPyBandits.GitHub.io/}} that encompasses a lot of MAB algorithms published until mid-2019.

Table \ref{tab:SummaryOSA-PoC_1} summarizes the main characteristics of the RL algorithm implementation for this use case. More details about these measurements can be found in \cite{Moy_2014}.

\begin{table}[htbp]
\centering 
\scalebox{0.85}
{\begin{tabular}{|c|c|c|c|c|}
\hline
Characteristics & Comment \\
\hline
\hline
RL algorithm &  UCB1 (or any other bandit algorithm)      \\
\hline
Reward & channel 'availability'      \\
\hline
Implementation side & secondary user   \\
\hline
method for RL feedback loop & Sensing   \\
\hline
\end{tabular}}
\caption{Summary for OSA proof-of-concept.}\label{tab:SummaryOSA-PoC_1}
\hspace{0in}
\end{table}

A PoC implementing MAB algorithms for internet of things (IoT) access has been done in \cite{Besson_2019}. It consists in one gateway, one or several learning IoT devices, embedding UCB1 and Thompson Sampling algorithms, and a traffic generator that emulates radio interferences from many other IoT devices. The IoT network access is modeled as a discrete sequential decision making problem. No specific IoT standard is implemented in order to stay agnostic to any specific IoT implementation. The PoC shows that intelligent IoT devices can improve their network access by using low complexity and decentralized algorithms which can be added in a straightforward and cost-less manner in any IoT network (such as Sigfox, LoRaWAN, etc.), without any modification at the network side. Table \ref{tab:SummaryIoT-PoC_1} summarizes the characteristics of the RL algorithm implementation for this use case. More details about these measurements can be found in \cite{Besson_2019}, but we present the main outcomes here.

\begin{figure}[htpb]
	\begin{center}
	\includegraphics[viewport = 200 90 520 410, clip, width=0.8\textwidth]{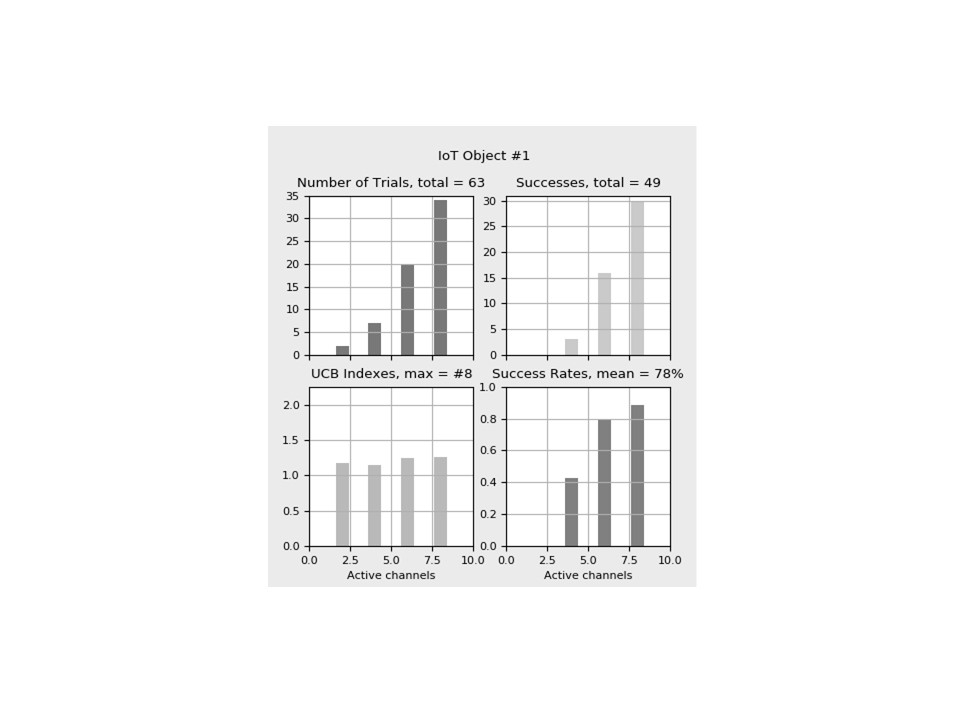}
	\caption{UCB parameters monitored during PoC execution \cite{Besson_2019}.}
	\label{fig:IoT-PoC_1}
	\end{center}
\end{figure}

Figure \ref{fig:IoT-PoC_1} shows the UCB parameters during an execution on 4 channels numbered as channels \#2, \#4, \#6 and \#8\footnote{The entire source code for this demo is available on-line, open-sourced under GPLv3 license, at \url{https://bitbucket.org/scee_ietr/malin-multi-armed-bandit-learning-for-iot-networks-with-grc/src/master/}. It contains both the GNU Radio Companion flowcharts and blocks, with ready-to-use Makefiles to easily compile, install and launch the demonstration.} \footnote{A 6-minute video showing the demonstration is at \url{https://www.youtube.com/watch?v=HospLNQhcMk&feature=youtu.be}}. We can see on top left of Fig. \ref{fig:IoT-PoC_1} that the left channel (channel $\#$2) has been only tried twice over 63 trials by the IoT device and it did not receive the ACK from the network at both times, as it can be seen at the top right panel which represents the number of successes on each channel. So the  success rate for channel $\#$2 is null, as seen at the bottom right panel. The more the channel index increases, i.e. from the left to the right, the better the success rate. That is why channel $\#$8, with 90\% of success rate has been preferably used by the algorithm with 35 trials over 63 (top left panel) and 30 successes (top right panel) over the 49 total successes obtained until the caption of this figure during the experiment.

\begin{table}[hbtp]
\centering 
\scalebox{0.85}
{\begin{tabular}{|c|c|c|c|c|}
\hline
Characteristics & Comment \\
\hline
\hline
RL algorithm &  UCB1 and Thomson Sampling      \\
\hline
Reward & channel 'availability'      \\
\hline
Implementation side & embedded on device   \\
\hline
method for RL feedback loop & Emulated ACK   \\
\hline
\end{tabular}}
\caption{Summary for IoT proof-of-concept.}\label{tab:SummaryIoT-PoC_1}
\hspace{0in}
\end{table}

\subsubsection{Real world experimentations}
The ultimate step of experimentation is the real world, validating the MAB approaches for intelligent spectrum access. It has been done on a LoRaWAN network deployed in the licence free 868 MHz ISM band for Europe. The experiment has been conducted in two steps. The first step consists of emulating an artificial IoT traffic in controlled radio conditions, i.e. inside an anechoic chamber, in order to validate the devices and the gateway implementation itself \cite{Moy_2020}. The second step is to make it run in real world conditions without being able to neither control the spectrum use, nor the propagation conditions in the area \cite{Moy_2019a}. On step 1, seven channels have been considered, whereas on step 2, only three channels were used due to the configuration of the gateway, which was controlled by the LoRaWAN network provider. For the two measurement campaigns, Pycom equipped with Lopy4\footnote{https://pycom.io/} shields have been used as devices and a standard  gateway from Mutlitech.

STEP 1 - 
The characteristics of the seven channels are given in Table \ref{tab:ChannelLoRaWAN_1} which gives the index of the channel, the percentage of time occupancy (or jamming) the channels experience due to other IoT devices in the area, the center frequency of each channel (channel bandwidth is set to 125 kHz). USRP platforms have been used to generate the surrounding IoT traffic.

\begin{table} 
\centering 
\scalebox{0.85}{\begin{tabular}{|c|c|c|c|}
\hline
Channel & \% of jamming  &  Frequency (in MHz) 
 \ \\
\hline 
$\#$0 & 30\% &  866.9       \    \\
\hline
$\#$1 & 25\% & 867.1        \    \\
\hline
$\#$2 & 20\% & 867.3        \    \\
\hline
$\#$3 & 15\% & 867.5        \    \\
\hline
$\#$4 & 10\% & 867.7        \    \\
\hline
$\#$5 & 5\%  & 867.9        \    \\
\hline
$\#$6 & 0\%  & 868.1        \    \\
\hline
\end{tabular}}
\caption{Channels characteristics for step 1 experiments.}\label{tab:ChannelLoRaWAN_1}
\hspace{0in}
\end{table}

\begin{figure}[htpb]
	\begin{center}
 	\includegraphics[viewport = 40 100 650 410, clip, width=\textwidth]{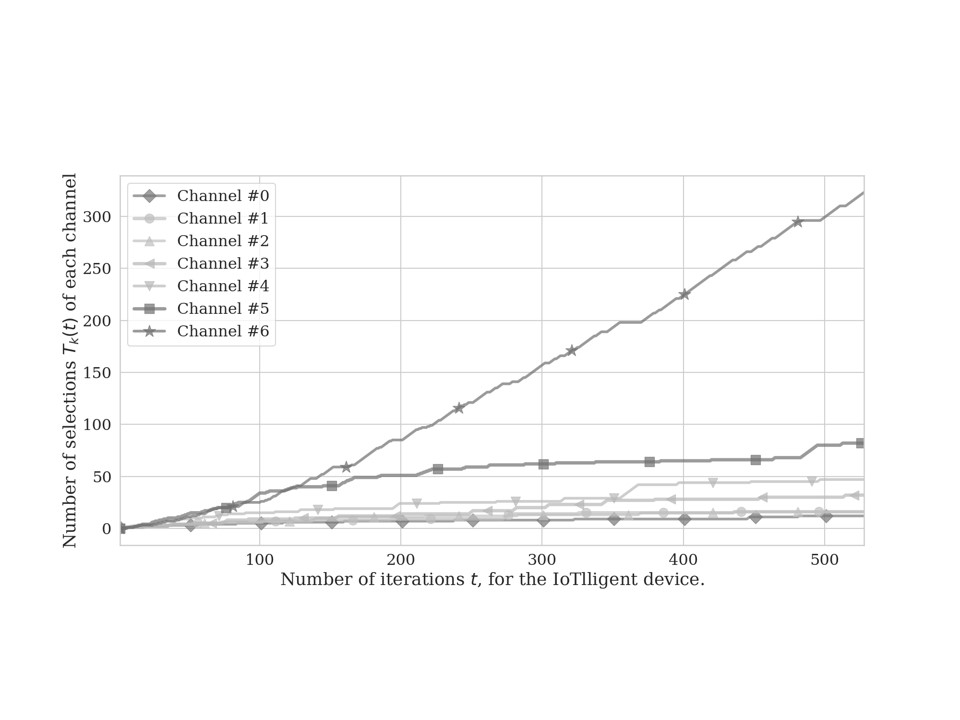}
	\caption{Number of selections for each of the seven channels used in step 1 for IoTligent device  \cite{Moy_2020}.} 
	\label{fig:LoRaWAN_1}
	\end{center}
\end{figure}

Figure \ref{fig:LoRaWAN_1} shows that, due to the surrounding IoT traffic, the channel number $\#$6, i.e. the curve with star markers, has been much more played by the cognitive device, thanks to the learning algorithm it embeds. This device is therefore named IoTligent and is hence able to maximise the success rate of its transmission. A transmission is called a "success" when a message has been sent by the IoT device in uplink, received by the LoRaWAN gateway, transmitted to the application server which sends an ACK back towards the IoT device, which is transmitted by the gateway in downlink at the same frequency used in uplink and finally received by the IoT device. A normal device following standard LoRaWAN features used by default, served as reference and the results in term of number of channel uses and success rate have been summarized in Table \ref{tab:ResultsLoRaWAN_1}. 

As we can see on Table \ref{tab:ResultsLoRaWAN_1}, whereas the reference IoT device uniformly transmits on all channels (around 75 times during this experiment), we can see that the IoTligent device concentrates its transmissions on the most vacant channels, with a clear choice for channel $\#$6. Over a total of 528 iterations, 323 transmissions have been done in this channel for IoTligent, which is more than 4 times compared to the reference IoT, with 75 transmissions. Moreover, IoTligent selects the channel $\#$6 much more than channel $\#$0, i.e. almost 27 times more. Hence, the IoT device with learning capability is able to increase its global success rate drastically that reaches 80\% (420 successful ACK received over 528) compared to 50\% for the reference IoT device (266 successful ACK received only). 

\begin{table} 
\centering 
\scalebox{0.85}
{\begin{tabular}{|c|c|c|c|c|}
\hline
 & Reference IoT & & IoTligent &     \\
\hline
Channel & \% of success  &  nb of Tx  & \% of success  &  nb of Tx
 \\
\hline 
$\#$0 & 21\%   &  76   &  8\%    &  12      \\
\hline
$\#$1 & 20\%   &  76   &  25\%   &  16      \\
\hline
$\#$2 & 24\%   &  75   &  25\%   &  16      \\
\hline
$\#$3 & 49\%   &  76   &  50\%   &  32      \\
\hline
$\#$4 & 62\%   &  74   &  62\%   &  47      \\
\hline
$\#$5 & 76\%   &  76   &  74\%   &  82      \\
\hline
$\#$6 & 96\%   &  75   &  94\%   &  323     \\
\hline
\end{tabular}}
\caption{Success rate and number of attempts for each channel of regular IoT device and IoTligent.}\label{tab:ResultsLoRaWAN_1}
\hspace{0in}
\end{table}
On this example, due to its ability to favor the use of less occupied channels, IoTligent demonstrates 2.5 times improvement in performance compared to standard IoT LoRaWAN device, in terms of number of successes. Note that in this experimental setup, radio collisions are only considered as obstacles to the reception of ACK by the IoT devices. 

It is worth mentioning that adding this learning capability does not impose changes to LoRaWAN protocols: no additional re-transmissions to be sent, no additional power consumed, no data to be added in frames. The only condition is that the proposed solution should work with the acknowledged (ACK) mode for IoT. The underlying hypothesis, however, is that the channels occupancy by surrounding radio signals (IoT or not) is not equally balanced. In other words, some ISM sub-bands are less occupied or jammed than others, but it is not possible to predict it in time and space, so the need to learn on the field. Table \ref{tab:SummaryLoRaWAN_1} sums-up the characteristics of the RL algorithm implementation for this use case. More details about these measurements can be found in \cite{Moy_2020}.

\begin{table}[htbp]
\centering 
\scalebox{0.85}
{\begin{tabular}{|c|c|c|c|c|}
\hline
Characteristics & Comment \\
\hline
\hline 
RL algorithm &  UCB1 (or any other bandit algorithm)      \\
\hline
Reward & received ACK      \\
\hline
Implementation side & embedded on device   \\
\hline
method for RL feedback loop & Standard LoRaWAN ACK   \\
\hline
\end{tabular}}
\caption{Summary for step 1 experiments}\label{tab:SummaryLoRaWAN_1}
\hspace{0in}
\end{table}

STEP 2 - Real world experiments have been done on a LoRa network deployed in the town of Rennes, France, with 3 channels {868.1 MHz, 868.3 MHz, 868.5 MHz}. IoTligent is completely agnostic to the number of channels and can be used in any country or ITU Region (i.e. 866 MHz and 915 MHz ISM bands as well). Since this experiment is run in real conditions, we have no means to determine exactly which of the four following possible phenomenon influences the IoTligent devices behavior:  i) collisions with other LoRaWAN IoT devices, ii) collisions with IoT devices running other IoT standards, iii) collisions with other radio jammers in the ISM band, iv) propagation issues.

\begin{table}[b]
\centering 
\scalebox{0.85}
{\begin{tabular}{|c|c|c|c|c|}
\hline
Characteristics & Comment \\
\hline
\hline 
RL algorithm &  UCB1 (but could be any bandit algorithm)      \\
\hline
Reward & channel 'availability' (collisions, jamming and propagation)      \\
\hline
Implementation side & embedded on device   \\
\hline
method for RL feedback loop & Standard LoRaWAN ACK   \\
\hline
\end{tabular}}
\caption{Summary for step 2 experiments.}\label{tab:SummaryLoRaWAN_2}
\hspace{0in}
\end{table}

\begin{figure}[hbtp]
	\begin{center}
	\includegraphics[viewport = 40 90 660 403, clip, width=\textwidth]{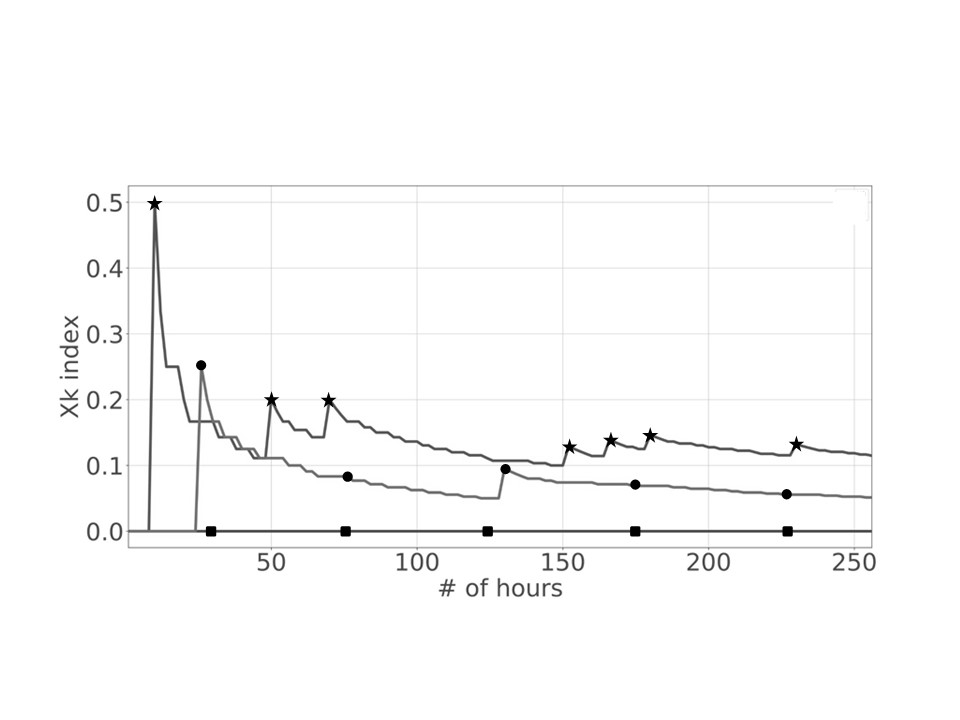}
	\caption{Empirical mean evolution through time over 11 days \cite{Moy_2019a}} 
	\label{fig:LoRaWAN2}
	\end{center}
\end{figure}

We now look at the results obtained by IoTligent, for 129 transmissions done every 2 hours, over an 11 days period. Figure \ref{fig:LoRaWAN2} shows the empirical mean experienced by the device on each of the 3 channels. This represents the average success rate achieved in each channel since the beginning of the experiment. The average success rates for the three channels, i.e. $\#$0 (868,1 MHz), $\#$1 (868.3 MHz) and $\#$2 (868.5 MHz), are represented by the curves with squares, stars and bullets, respectively. Each peak corresponds to a LoRa successful bi-directional exchange between the device and the application server: from device transmission, to ACK reception by the device. Each peak in Figure \ref{fig:LoRaWAN2} reveals a successful transmission where ACK has been received by IoTligent device. We can see that channel $\#$1, star markers, has been the most successful, before channel $\#$2, while channel $\#$0 always failed to send back an ACK to the device. During the experiment indeed, channel $\#$0 has been tried 29 times with 0 success. IoTligent device uses channel $\#$1 61 times with 7 successful bi-directional exchanges, i.e. 7 peaks on the curve with stars, and channel $\#$2 39 times  with 2 successes, i.e. 2 peaks on the curve with bullets. At the end of the experiment, one can observe 11.5\% successful bi-directional connections for channel $\#$1 and 5\% for channel $\#$2, whereas channel $\#$0 never worked from the device point of view. For comparison, a regular IoT device, performing a random access, achieves a global average successful rate of 5.5\%. Table \ref{tab:SummaryLoRaWAN_2} summarizes the characteristics of the RL algorithm implementation for this use case. More details about these measurements can be found in \cite{Moy_2019a}.

\section{Conclusions and Future Trends} \label{sec:trends}

During recent years, we have witnessed a shift in the way wireless communication networks are designed, at least in the academic community, by introducing ML techniques to optimize the networks. With the complexity of wireless systems increasing, it becomes more and more difficult to build explainable mathematical model to predict the performance of large scale systems. Exhaustive and complex simulations are not always an option because they are highly resource-intensive. In this sense, ML in general and RL in particular have the potential to overcome the network modelling deficit by learning the optimal network functioning with minimal assumptions on the underlying phenomena. This is of particular importance when non linear phenomena needs to be taken into account and for which it is very difficult to get analytical insights.

In this Chapter, we focused on RL for PHY layer communications. Even if this domain enjoys a long successful history of well-defined mathematical models, the heterogeneity of the use-cases envisaged in 5G for instance, advocates for adding smartness in the PHY layer of upcoming wireless systems. All along this Chapter, we have provided some PHY examples in which RL or deep RL can help to achieve good performance results. The key point to apply RL algorithms to a PHY layer problem is when one cannot solve the problem analytically or by writing an \emph{explicit program} to do so. There are, without any doubt, many practical situations where one cannot \emph{explicitly} solve the associated optimization problem related to the communication system design, e.g. non linear peak to average power ratio reduction, acoustic transmission, non linear power-amplifier.

The key benefit of RL is its adaptability to unknown system dynamics. However, the convergence time is crucial in wireless communications and real networks cannot afford to spend too much time to learn a good strategy. For instance, the deep Q-learning in \cite{Sadeghi2018} converges around $10^4$ iterations, for two popularity profile states. This cannot be directly converted into time delay because it depends on the number of requests the users do, but if the network records 1 request per second, it takes almost 3 hours to converge toward the optimal strategy. The converging time is also around $10^3$-$10^4$ iterations in the problem of optimal buffer state and transmission power dealt with in \cite{Mastronarde2013}. In case of non stationary environment, it can be prohibitive.

The feasibility of a PHY layer data-driven designed has been proved in \cite{Shea2017,Dorner2018}. However, even if possible, learning an entire PHY layer without any \emph{expert knowledge} is not necessarily desirable. For instance, the synchronisation task requires an huge amount of computation and a dedicated NN in \cite{Dorner2018} to synchronize the transmission while it can be simply performed using well-known OFDM signal structures \cite{Felix2018}. Hence it is apparent that DRL can be an efficient tool to design the PHY layer of future wireless systems, however, it would may gain in efficiency if cross-fertilization research between ML and model-based approach would be undertaken. In order to achieve this goal, explainable ANN with information-theoretic tools is a promising research direction, as attempted in \cite{Louart2018}. The \emph{transfer learning} concept, which consists in transferring the knowledge acquired in a previous task to a current one in order to speed up the converge, is also a promising research direction if a mixed transfer, i.e. model-based to data-driven, is considered.
\section{Bibliographical remarks} \label{sec:relatedworks}

This Chapter aimed at being the most self-content as possible and at presenting, in a tutorial way, the theoretical background of RL and some application examples, drawn from the PHY layer communication domain, where RL techniques can be applied. The theoretical part of Section \ref{sec:rltheory} revisits the fundamentals of RL and is largely based on the great books of Sutton and Barto \cite{Sutton2018} and Csaba Szepesvari \cite{Szepesvari2010} that gathered in a comprehensive way the original results of Bellman \cite{Bellman1957} who introduced the notion of dynamic programming and Watkins \cite{Watkins1989} who proposed the Q-learning algorithm. 

Example in Section \ref{sec:qlearex} follows from \cite{Ngo2010,Mastronarde2013}. The problem of joint power and delay management can be dated back to at least the early 2000s. In \cite{Berry2002} the authors studied the power adaptation and transmission rate to optimize the power consumption and the buffer occupancy as well. They characterized the optimal power-delay tradeoff and provided the former steps of a dynamic scheduling strategy achieving the Pareto front derived earlier. While the previous work investigated the power-delay Pareto front in block fading channels, authors in \cite{Rajan2004} came back to AWGN channel but proposed more explicit schedulers to operate close to the fundamental bound. The authors in \cite{Goyal2003} revisited the same problem as in \cite{Berry2002} and proved the existence of a stationary policy to achieve the power-delay Pareto front. The formulation of the joint delay-power management problem as a constrained MDP can be reported in \cite{Goyal2003,Salodkar2008,Ngo2010,Mastronarde2013}.

Example in Section \ref{sec:linearrl} has been considered in \cite{Sadeghi2018} with a more complex setting, where the authors introduced two kind of popularity profiles, i.e. local and global. The local popularity profile allows to cache the requested files according to the local demand while the global one allows to anticipate the local demand by monitoring the most wanted files over the network. We simplified the system model in order to make a toy example easily with this application. Due to the potentially large number of files to be cached, this problem can be addressed with deep neural network as well, see for instance \cite{Zhong2020,He2017}.

Examples of Section \ref{sec:mabex} follow from \cite{Oksanen2015,Navik_2017} for the OSA problem with quality of the transmission and from \cite{Navik2019} for the switch ON/OFF base station problem. The opportunistic spectrum access problem has been considered as the typical use-case for the application of MAB and RL algorithms in wireless communications since their inception in the field. This matter of fact comes from the RL framework in general, and MAB in particular, are well suited for describing the success and the failure of a user that tries to opportunistically access to a spectral resource that is sporadically occupied. The second reason is that the performance of the learning algorithm strongly depends on the ability of the device to detect good channels, that motivates the reborn of the research on signal detection algorithms. A broad overview on signal detection and algorithms to opportunistically exploit the frequency resource has been provided in \cite{Lunden_2015}.

The examples we proposed in this chapter are not exhaustive since it was not the objective and at least two very complete surveys have been provided recently on that topic \cite{Mao2018,Luong2019}. We invite the interested reader to consult these articles and the references therein to go further in the application of (deep)-RL techniques to some specific wireless communication problems. In this section, we give some articles that may be interesting to consult when addressing PHY layer communication challenges.

An interesting use-case in which DRL can be applied is the IoT. The huge number of cheap devices to be connected with small signalling overheads make this application appealing for some learning approaches implemented at BS for instance. Authors in \cite{Wang2018} revisited the problem of dynamic spectrum access for one user. Even if this set up has been widely studied in literature, the authors modeled the dynamic access issue as a POMDP with correlated states and they introduce a deep Q-learning algorithm to choose what are the best channels to access at each time slot. An extension to multiple devices has been provided in \cite{Zhu2018}. A relay with buffering capability is considered to forward the packets of the other nodes to the sink. At the beginning of each frame, the relay chooses packets from buffers to transmit on some channels with the suitable power and rate. They propose a deep Q-learning approach with a FNN to optimize the packet transmission rate. IoT networks with energy harvesting capability can also be addressed with deep Q-learning in order to accurately predict the battery state of each sensor \cite{Chu2019}.

The works above consider centralized resource allocation, i.e. the agent is run in the BS, relay or another server but is located at one place. In large scale heterogeneous networks, with multiple kind of BS, e.g. macro, small or pico BS, this approach is no longer valid and distributed learning has to be implemented. In \cite{Challita2018}, the authors considered the problem of LTE access through WiFi small cells by allocating the communication channels and managing the interference. The small BS are in competition to access the resources and hence the problem is formulated as a non cooperative game which is solved using deep RL techniques. The same kind of problem has been tackled in \cite{Naparstek2019} while in a different context and different utility function.

DRL has also been successfully applied to complex and changing radio environments with multiple conflicting metrics such as in satellite communications in \cite{Ferreira2018}. Indeed, the orbital dynamics, the variable propagation environment, e.g. cloudy, clear sky, the multiple optimization objectives to handle, e.g. low bit error rate, throughput improvement, power and spectral efficiencies, make the analytical optimization of the global system untractable. Authors used a deep Q-learning with a deep NN to choose the actions to perform at each cognitive cycle, e.g. modulation and encoding rate, power transmission, and demonstrate that the proposed solution achieved very good performance compared to the ideal case obtained with a brute force search.

There are still lots of papers dealing with RL and deep-RL for PHY layer communications, ranging from resource allocations to PHY layer security for instance. Since this is a hot topic rapidly evolving at the time we write this book, one can expect very interesting and important contributions in this field in the upcoming years. Since the feasibility and the potential of RL techniques has been demonstrated for PHY layer communications, we encourage the research community to also address important issues such as the convergence time reduction of learning algorithms or the energy consumption reduction for training a deep NN instead of an expert-based design of PHY layer.

\section*{Acronyms}
\addcontentsline{toc}{section}{Acronyms}
\vskip 1em
\begin{table}[htbp]
\renewcommand{\arraystretch}{1.25}
\begin{tabular}[12pt]{p{3cm}p{10cm}}
ACK & Acknowledgement \\
ANN & Artificial Neural Network \\
AWGN & Additive White Gaussian Noise \\
BPSK & Binary Phase Shift Keying \\
BS & Base Station \\
DL & Deep Learning \\
DRL & Reinforcement Deep Learning \\
HF & High Frequency \\
HMM & Hidden Markov Model \\
MAB & Multi-Armed Bandit \\
MDP & Markov Decision Process \\
NN & Neural Network \\
OFDM & Orthogonal Frequency Division Multiple Access \\
PHY & Physical layer of the OSI stack \\
POMDP & Partially Observable Markov Decision Process \\
ML  & Machine Learning \\
QoS & Quality of Service \\
RL & Reinforcement Learning \\
SINR & Signal to Interference plus Noise Ratio \\
SU & Secondary User(s) \\
UCB & Upper Confidence Bound \\
 \end{tabular}
\end{table}
\clearpage

\section*{Notation and symbols}
\addcontentsline{toc}{section}{Notation and symbols}
\begin{table}[htbp]
\renewcommand{\arraystretch}{1.25}
\begin{tabular}{ll}
$A(t)$ & Action random variable at $t$ \\
$a$ & A realization of the random variable action $A$ \\
$\Exp{X(t)}{P}$ & Expectation of $X(t)$ under the distribution $P$\\
$\prob{X(t) = x}{X}$ & Probability measure of the event $X(t)=x$, under the distribution of X \\
$P_{X(t+1)\left|\right.X(t)}(x'\left|\right. x)$ & Conditional probability measure of $X$ at $t+1$ knowing $X$ at $t$ \\
$\pi(a\left|\right. s)$ & Policy $\pi$, i.e. probability to choose action $a$, while observing state s \\
$\pi^*$ & Optimal policy \\ 
$q_{\pi}(a,s)$ & Action-state value of the pair $(a,s)$ and following hereafter the policy $\pi$ \\
$q_*(a,s)$ & Optimal action-state value of the pair $(a,s)$ and following hereafter $\pi^*$ \\
$R(t)$ & Reward random variable at $t$ \\
$r$ & A realization of the random variable reward $R$ \\
$S(t)$ & State random variable at $t$ \\
$s$ & A realization of the random variable state $S$ \\
$v_{\pi}(s)$ & Value function of the state $s$ under the policy $\pi$ \\
$v_*(s)$ & Optimal value function of the state $s$ under $\pi^*$\\
& \\
$\left[\cdot\right]^+$ & $\max\left(\cdot,0\right)$ \\
$\mathds{1}\left\{a\right\}$ & Indicator function equals to 1 if $a$ is true and 0 otherwise \\
$\vec{1}_n$ & Column vector full of ones of length $n$ \\
& \\
$\set{A}$ & Set of possible actions \\
$\mathbb{N}$ & Set of positive integers \\
$\set{R}$ & Set of possible rewards \\
$\mathbb{R}$ & Set of real numbers \\
$\mathbb{R}_+$ & Set of real and positive numbers \\
$\set{S}$ & Set of possible states \\
\end{tabular}
\end{table}

\clearpage

\setcounter{tocdepth}{2}
\tableofcontents
\clearpage

\bibliographystyle{IEEEtran}
\bibliography{references}
\end{document}